\documentclass[10pt,twocolumn,letterpaper]{article}

\usepackage[pagenumbers]{cvpr}

\newcommand{\gain}[1]{{\scriptsize\textcolor{gray}{(#1)}}}

\usepackage{adjustbox}
\newsavebox{\fullpagecapbox}

\definecolor{cvprblue}{rgb}{0.21,0.49,0.74}
\usepackage[pagebackref,breaklinks,colorlinks,allcolors=cvprblue]{hyperref}
\usepackage{multirow}

\def\paperID{*****}
\def\confName{CVPR}
\def\confYear{2027}

\title{FlowBench: Counterfactual Imagination Training for Spatial Intelligence}

\author{Yifan Liu$^*$, Baihan Zhang$^*$, Khai Loong Aw$^*$, Klemen Kotar$^*$, Daniel L.K. Yamins \\ Stanford University
\\ \small{$^*$Equal contribution}}

\begin{document}
\maketitle
\begin{abstract}

Humans imagine scene changes under hypothetical motions, an ability central to spatial reasoning and robotic planning. However, this capability remains underexplored in vision-language models and world models. We formalize this capability as Counterfactual Spatial Imagination (CSI) and introduce FlowBench, a resource for training and evaluating both, with 428K training examples spanning simulated and real-world scenes, household environments and tabletop manipulation, and diverse question types. A scalable, fully automatic pipeline pairs each hypothetical motion with spatial questions and, in simulation, exact outcome frames and optical flow. For VLMs, proprietary models exhibit an imagination gap, performing worse when motion outcomes must be inferred than when observed, and open-source models perform near chance; fine-tuning Qwen3-VL-8B on FlowBench-Sim-Train raises its accuracy by 29.5 percentage points in simulation and 16.9 on real images, a strong sim-to-real transfer. For world models, training on FlowBench-Sim improves text-to-flow prediction for all four model families we test, raising the accuracy of $\Psi$-0.6 from 52.1\% to 61.7\%. FlowBench thus serves as a training and evaluation resource for both. Code and datasets are released at \url{https://github.com/liuyifan22/FlowBench}.

\end{abstract}

\section{Introduction}
\begin{figure}[t]
  \centering
  \includegraphics[width=1.0\columnwidth]{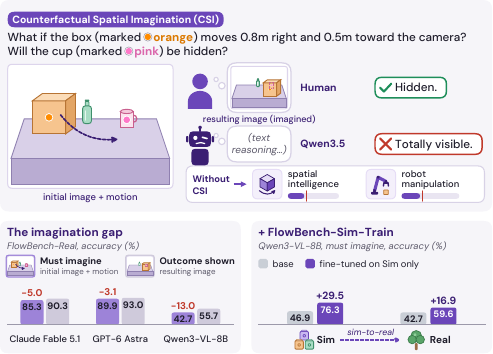}
\caption{\textbf{Counterfactual Spatial Imagination (CSI) is a core capability for acting in the world.}
Given the initial image and a hypothetical motion, a human imagines the outcome, whereas open-source VLMs
such as Qwen3.5 reason in text and fail (top).
\textbf{Even proprietary models show an imagination gap} (bottom left): on FlowBench-Real-Test,
Claude-Fable-5.1 and GPT-6~Astra lose 5.0 and 3.1 points when they must imagine the outcome instead of
seeing it; Qwen3-VL-8B loses 13.0.
\textbf{Training on FlowBench-Sim transfers to real scenes} (bottom right): fine-tuning Qwen3-VL-8B on
FlowBench-Sim-Train alone adds 29.5 points on Sim and 16.9 on Real.}
  \label{fig:teaser}
\end{figure}

Humans can mentally simulate scene changes under hypothetical interventions~\citep{tolman1948cognitive,shepard1971mental,hassabis2007patients}.
For example, we can anticipate whether moving a sofa would block a corridor without physically moving it.
This spatial \textit{foresight} supports interaction with the physical world and could help robots anticipate how object motions affect free space, visibility, and proximity during navigation or manipulation.

Recent vision-language models (VLMs) have demonstrated strong visual perception and reasoning abilities~\citep{achiam2023gpt,comanici2025gemini,bai2025qwen3}, but understanding an observed scene does not necessarily imply anticipating its state after a hypothetical motion. Existing spatial benchmarks primarily evaluate observed visual information~\citep{yin2025mindcube,zhang2026theory,hong2026esi}, while work on mental simulation often considers camera motion in static scenes~\citep{ray2024sat,yang2026mindjourney,liu2026abstract}.
Current counterfactual visual question answering benchmarks~\citep{niu2021counterfactual,chen2025countervqa,cai2026if} focus on semantic changes, leaving open whether VLMs can infer the spatial consequences of object motions in realistic scenes.
These directions leave a gap in evaluating changes to the spatial configuration itself, beyond interpreting an observed scene or viewing it from another perspective.
World models offer a complementary route: rather than reasoning in text, they simulate a described motion visually, predicting its optical flow or generating the frame after it~\citep{psi,ran_fofpred_2026,brooks2023instructpix2pix}, and their predictions are increasingly evaluated for physical consistency under interventions~\citep{cai2026if}. Training them for this requires text commands paired with the exact flow and outcome of the motion they describe, whereas video corpora offer only estimated flow and motion descriptions written after the fact.
We therefore ask:

\textit{How well can VLMs and world models reason about and simulate how a scene changes under hypothetical object motions, and can targeted training improve both?}

We formalize this capability as \textit{Counterfactual Spatial Imagination} (CSI).
Given a scene image, a hypothetical object motion, and a spatial question, a model must answer with respect to the post-motion scene without observing it.
CSI requires understanding the initial scene, interpreting the motion, and inferring its consequences for the relevant spatial relationships.
We evaluate answers without requiring intermediate image generation, accommodating both implicit reasoning in VLMs and explicit visual simulation with world models; for world models, we additionally evaluate the simulation itself through its predicted optical flow and outcome frame.

We introduce \textit{FlowBench}, a training dataset and benchmark with approximately 428K training examples and held-out test splits spanning simulated household environments and real tabletop manipulation.
FlowBench-Sim covers eight question categories probing changes in spatial relationships, visibility, and collision constraints relevant to navigation through changing layouts.
FlowBench-Real focuses on object rearrangements across seven question categories.
Their test splits assess generalization to unseen simulated scene styles and sim-to-real transfer.

We develop a scalable, fully automatic pipeline for accurate CSI data generation that others can adapt to new settings.
For FlowBench-Sim, deterministic programs compute ground-truth answers from AI2-THOR and ProcTHOR simulator metadata~\citep{ai2thor,procthor}.
The pipeline scales across procedurally generated scenes and balances samples across question categories.
Beyond question-answer pairs, each simulated motion comes with its text command, post-motion image, optical flow, depth, and semantic segmentation, all exact from simulator metadata, as well as textual reasoning annotations. The same data thus supervises VLMs, through questions and reasoning traces, and world models, through text-to-flow prediction and text-conditioned generation of the outcome frame.
For FlowBench-Real, we extract motions from recorded robot videos, relabel them as hypothetical motion, and compute answers from estimated post-motion geometry and visibility.
Evaluation examples undergo manual quality review, while simulated training data undergo VLM-assisted plausibility and correctness filtering.

Our experiments reveal that proprietary VLMs retain an \textit{imagination gap}: they perform worse when motion outcomes must be inferred than when those outcomes are directly observed.
As illustrated in \cref{fig:teaser}, Claude-Fable-5.1 and GPT-6 Astra exhibit gaps of 5.0 and 3.1 percentage points, respectively.
Off-the-shelf open-source VLMs perform near chance and below the human baseline, further motivating the need for stronger counterfactual world-modeling capabilities in VLMs.

Fine-tuning Qwen3-VL-8B on FlowBench-Sim-Train improves performance by 29.5 percentage points on FlowBench-Sim-Test and 16.9 points on FlowBench-Real-Test, demonstrating strong sim-to-real transfer of the proposed data annotation pipeline with simulators, showing potential of its application to other tasks.
Nevertheless, its imagination gap on real images persists after fine-tuning, while
text-based chain-of-thought reasoning also offers inconsistent gains.

On the world-model side, training on FlowBench-Sim-Train improves text-to-flow prediction on held-out scenes for all four model families we test, by up to 25.2 percentage points; the trained $\Psi$-0.6~\citep{psi} reaches 61.7\%, above a modular perception pipeline given camera intrinsics (\cref{sec:exp_wm_train}). Letting $\Psi$-0.6 imagine the outcome frame before answering does not yet help, as generated frames lose details the questions depend on, so world models must become more faithful before they can support VLM reasoning.

Our contributions are fivefold:
\begin{itemize}
    \item We formalize CSI as an important spatial reasoning capability for anticipating scene changes under hypothetical object motions.
    \item We introduce FlowBench, a diverse training dataset and benchmark spanning simulated household environments, real tabletop manipulation, and a broad range of spatial question categories.
    \item We develop a scalable, fully automatic pipeline for accurate dataset construction that pairs each motion with spatial questions, reasoning traces, and exact outcome supervision (post-motion image, optical flow, depth, and segmentation), supporting the training of both VLMs and world models.
    \item For VLMs, we show that proprietary VLMs can exceed human baseline on FlowBench-Real yet retain an imagination gap, while open-source models are near chance and benefit from targeted training with strong sim-to-real transfer.
    \item For world models, we show that training on FlowBench-Sim improves text-to-flow prediction for every model family we test, while world-model-assisted answering does not yet help CSI.
\end{itemize}
Our code and datasets are publicly released to support research on CSI in both VLMs and world models, and on spatial foresight for robotic navigation and manipulation.
\begin{figure*}[t]
  \centering
  \includegraphics[width=1.0\textwidth]{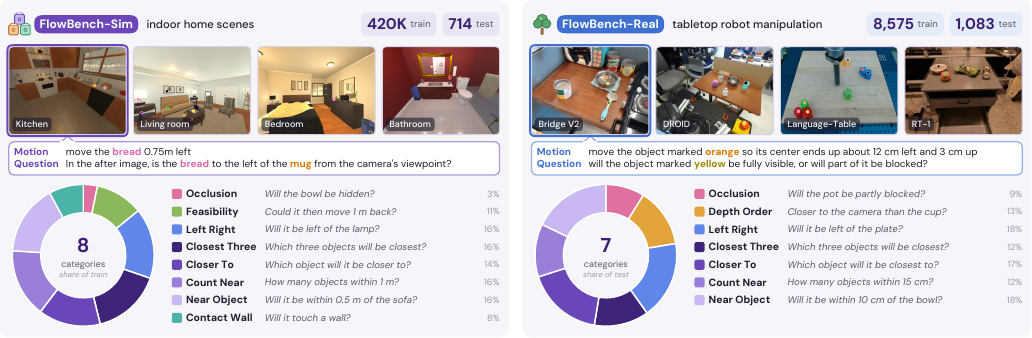}
\caption{\textbf{FlowBench is diverse and comprehensive: simulated homes and real robot tabletops, nine question categories, and 428K training questions.}
FlowBench-Sim covers kitchens, living rooms, bedrooms, and bathrooms in about 10K procedurally generated
houses (420K training questions; 714 test questions from 120 held-out AI2-THOR scenes). FlowBench-Real
covers tabletop manipulation from five robot datasets (8,575 training and 1,083 test questions).
Questions probe visibility, relative position, proximity, depth order, feasibility, and contact: eight
categories in Sim and seven in Real, six of them shared (shares of the training and test split,
respectively).}
  \label{fig:dataset_overview}
\end{figure*}

\section{Related Work}

\noindent\textbf{VLMs and Spatial Intelligence.}
Spatial intelligence involves understanding and reasoning about the spatial structure of the physical world.
VLM research examines spatial perception, memory, and recall~\citep{yang2025thinking}, alongside cognitive mapping~\citep{yin2025mindcube}.
Further work studies the active construction and revision of spatial beliefs or abstractions~\citep{zhang2026theory,yang2026mindjourney,liu2026abstract} and 3D-based information aggregation~\citep{hong20233d,lin2026qwen}.
Related benchmarks study motion perception as part of physical understanding~\citep{riochet2018intphys,bakhtin2019phyre,bordes2025intphys2,patel2022cripp}.
Spatial instruction datasets and spatially grounded supervision also improve VLM spatial reasoning~\citep{ray2024sat,chen2024spatialvlm,cheng2024spatialrgpt,cai2026sensenova}.
However, these studies primarily address perception, memory, and reasoning about observed environments.
Humans can also imagine alternative spatial configurations and reason about them~\citep{kosslyn1980image,epstein1998cortical,gardner2011frames}; developing this counterfactual spatial imagination in VLMs remains an open challenge.

\noindent\textbf{Counterfactual Reasoning.}
Counterfactual reasoning infers the consequences of hypothetical changes to observed reality.
Humans naturally construct counterfactual mental models for learning, planning, and prediction~\citep{kahneman1982simulation,byrne2002mental,van2015cognitive}.
From a causal perspective, this requires understanding how interventions alter outcomes rather than relying on observational correlations~\citep{pearl2009causality}.
Multimodal studies examine language bias under hypothetical visual changes~\citep{niu2021counterfactual}, as well as visual causal reasoning and video understanding~\citep{komanduri2025causalvlbench,chen2025countervqa}.
Beyond question answering, generative world models are evaluated on physically consistent future prediction under controlled interventions, advancing toward causal simulation of embodied environments~\citep{cai2026if}.
However, existing benchmarks primarily address semantic changes, causal attribution, or future visual generation; predicting how spatial relationships evolve under hypothetical interventions remains underexplored in VLMs.

\section{Method}

\subsection{Overview}
FlowBench evaluates Counterfactual Spatial Imagination (CSI) through multiple-choice visual question answering~\citep{antol2015vqa}.
Each example consists of
\begin{equation}
    (I_0, t_m, t_q, t_a),
    \label{eq:necessary}
\end{equation}
where $I_0$ is the initial scene image, $t_m$ describes a hypothetical object motion, $t_q$ is a spatial question with answer choices, and $t_a$ is the answer for the resulting scene.
The model predicts $t_a$ from $(I_0,t_m,t_q)$ without observing the outcome.
FlowBench combines simulated indoor environments and real robot manipulation scenes, targeting spatial reasoning relevant to navigation and manipulation.

\noindent\textbf{FlowBench-Sim.}
As shown in~\cref{fig:dataset_overview}, FlowBench-Sim-Train contains 420K questions from 10K ProcTHOR scenes~\citep{procthor}, covering eight question categories.
FlowBench-Sim-Test contains 714 questions from 120 AI2-THOR scenes~\citep{ai2thor} held out from the training scene source, evaluating generalization to unseen scene styles.

\noindent\textbf{FlowBench-Real.}
\label{sec:real_overview}
FlowBench-Real-Train contains 8{,}575 questions, and FlowBench-Real-Test contains 1{,}083 questions on separate image pairs.
Both are constructed from recorded robot manipulation videos: observed object displacements become hypothetical motions, while later frames provide the outcomes used for annotation.

\subsection{Benchmark Composition}
\label{sec:benchmark_contents}

\noindent\textbf{Question categories.}
The eight FlowBench-Sim training categories are Occlusion, Feasibility, Left Right, Closest Three, Closer To, Count Near, Threshold, and Contact Wall.
They cover visibility, the feasibility of subsequent motions, relative position, proximity, and wall contact.

\label{sec:real_composition}
FlowBench-Real shares six categories and adds Depth Order, which asks whether the moved object will be closer to the camera than a reference object.
Objects are identified by image marks~\citep{yang2023setofmark} or names; marked examples support both numbered marks and the colored dots used in simulation.
Each real example also has a static counterpart that asks the same question about the outcome image without a hypothetical motion.

\noindent\textbf{Auxiliary annotations.}
For simulated examples, we additionally provide richer annotations
\begin{equation}
    \{I_1,D_0,D_1,S_0,S_1,F_{01},
      c_{\mathrm{gt}},c_0,c_1\},
    \label{eq:auxiliary}
\end{equation}
where $I_1$ is the outcome image, $D_i$ and $S_i$ are depth and semantic segmentation, and $F_{01}$ is optical flow.
The three textual reasoning annotations are derived by VLM from simulator metadata ($c_{\mathrm{gt}}$), the outcome image ($c_1$), and the initial image with the motion ($c_0$).

\begin{figure*}[t]
  \centering
  \includegraphics[width=1.0\textwidth]{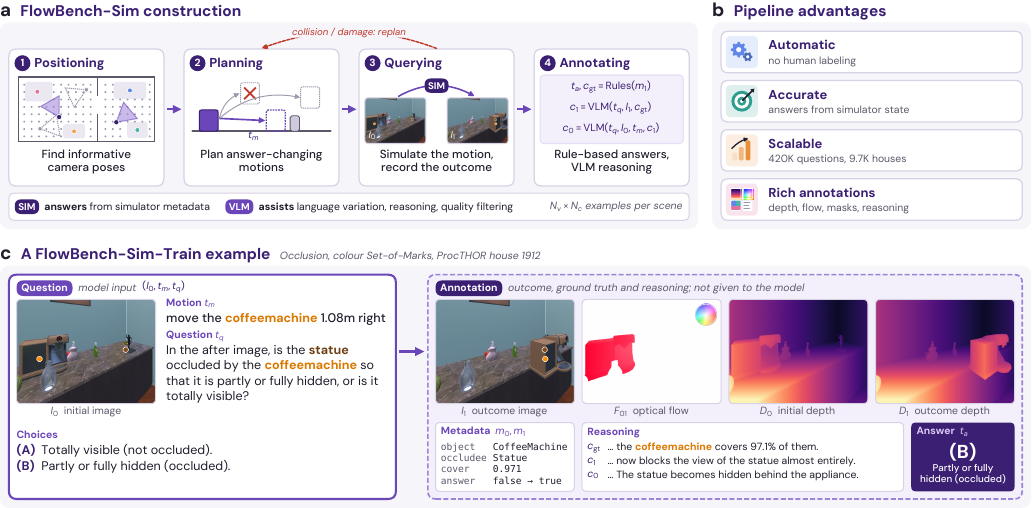}
  \caption{\textbf{A scalable, fully automatic, and accurate pipeline for counterfactual data that others can adapt to new simulators and tasks.}
\textbf{(a)}~Four stages: find informative views, plan answer-changing motions, execute them in the
simulator, and annotate. Answers are computed by rules on simulator metadata, so they are exact; a VLM
only adds reasoning and language variation.
\textbf{(b)}~The pipeline needs no human labeling, scales to 420K questions across 9.7K houses, and
records depth, flow, masks, and reasoning for every example.
\textbf{(c)}~A training example. The model sees only $(I_0,t_m,t_q)$; coloured nouns refer to the
marks of the same colour. The outcome image, flow, depth, metadata, and the reasoning chain
$c_{\mathrm{gt}}\!\rightarrow\!c_1\!\rightarrow\!c_0$ (excerpts) are training annotations that
supervise both VLMs and world models.}
  \label{fig:pipeline}
\end{figure*}

\subsection{Dataset Construction}
\label{sec:dataset_construction}

\subsubsection{FlowBench-Sim}
Our automated pipeline uses AI2-THOR simulator~\citep{ai2thor,procthor} to generate balanced CSI examples through four stages: Positioning, Planning, Querying, and Annotating.
Ground-truth answers are computed from simulator metadata; a VLM assists with language variation, reasoning annotations, and quality filtering.
For each scene, we target $N_vN_c$ examples across $N_v$ viewpoints and $N_c$ question categories.

\noindent\textbf{Positioning.}
We sample reachable camera poses and select views that balance movable-object visibility, spatial coverage, and viewpoint diversity.
(Suppl.~\ref{app:camera_selection}).

\noindent\textbf{Planning.}
Category-specific programs propose target configurations and derive the corresponding object motions.
This design balances answer options and favors motions that change the answer, reducing shortcuts based on the initial scene alone.
Infeasible proposals are resampled, and labels always reflect the realized outcome rather than the planned target
(Suppl.~\ref{app:planning_algorithms}).

\noindent\textbf{Querying.}
Given an initial scene state $\mathcal{S}_0$ and motion $t_m$, the simulator returns
\begin{equation}
    \mathcal{S}_1,m_{01}
    = \mathrm{SIM}(\mathcal{S}_0,t_m),
\end{equation}
where $m_{01}$ records transition information.

\noindent\textbf{Annotating.}
Deterministic, category-specific rules compute $t_a$ from the outcome metadata $m_1$, discarding ambiguous cases.
These rules also produce $c_{\mathrm{gt}}$, which fuels two VLM-generated reasoning annotations:
\begin{equation}
\begin{aligned}
    c_1 &= \mathrm{VLM}(t_q,I_1,c_{\mathrm{gt}}),\\
    c_0 &= \mathrm{VLM}(t_q,I_0,t_m,c_1).
\end{aligned}
\end{equation}
This hierarchy transfers supervision from simulator metadata to reasoning based on outcome images and then initial images with hypothetical motions.
Optical flow is computed from object metadata and depth. Question is answered based on rules; VLMs are only used for rephrasing.

\subsubsection{FlowBench-Real}
\label{sec:real}
FlowBench-Real measures object motions already in robot videos and recasts them as hypothetical interventions.

\noindent\textbf{Sources and frames.}
We use Bridge V2~\citep{walke2023bridgedata}, DROID~\citep{khazatsky2024droid}, Language-Table~\citep{lynch2023interactive}, RT-1~\citep{brohan2023rt1}, and Berkeley UR5~\citep{oxe2024}.
The initial frame precedes the action, and the outcome frame is the final frame or, for DROID, a suitable frame after the last gripper release. (Suppl.~\ref{app:real_mining}).

\noindent\textbf{Motion measurement.}
SAM~3~\citep{carion2025sam3} segments objects, Depth Anything~3~\citep{lin2025depthanything3} jointly estimates depth for the frame pair, and AllTracker~\citep{harley2025alltracker} provides trajectories for cross-frame matching.
Object positions are estimated by lifting mask centers into 3D using median mask depth. For selection, we kindly refer readers to Suppl.~\ref{app:real_measure}.

\noindent\textbf{Scale and motion descriptions.}
Motions are described as camera-relative horizontal and vertical displacements in centimeters, accompanied by a visible landmark's width as a scale reference.
We reject examples with inconsistent motion and reference scales. (Suppl.~\ref{app:real_scale}).

\noindent\textbf{Questions and answers.}
Answers are computed from estimated post-motion geometry and visibility, using stationary objects as landmarks.
All decisions require margins against measurement noise.
Most questions change their answer after the motion, while 291 (27\%) of FlowBench-Real-Test questions retain it to discourage shortcuts.
Questions are templated, and Qwen3.5-9B~\citep{qwen35} is used only to name objects in named-object examples.

\section{Experiments}
\label{sec:experiments}

\begin{table*}[p]
\sbox{\fullpagecapbox}{\parbox{\textwidth}{\caption{Accuracy (\%) in setting 1A on FlowBench-Sim (top) and FlowBench-Real (bottom): the model sees the \emph{before} image with colour Set-of-Marks and the action, and must answer about the scene after the action without seeing it. Wall and Feasible appear only in Sim and Depth only in Real (--).
Mean over three answer-order shuffles. Avg is per question; Cat-avg is the mean of the category accuracies. Bold: best per column within each group of models (proprietary, open-source, fine-tuned) and benchmark; chance and human rows are not bolded. The last rows of each block show models fine-tuned on FlowBench-Sim-Train; on FlowBench-Real they measure sim-to-real transfer. $^\ddagger$Pooled accuracy of human participants on a subset of each benchmark: 120 answers on FlowBench-Sim and 403 answers on FlowBench-Real (\cref{sec:exp_setup}). $^\dagger$Imagine, then answer: PSI-0.6 answers after generating the optical flow and the outcome frame (\cref{sec:analysis_wm}). $^\P$Evaluated on 50\% and $^\S$on 20\% of the questions of each benchmark. Low and max are the lowest and highest reasoning effort of the API (max reasons before answering, as in 1B). }
\label{tab:results_1a}}}%
\centering
\begin{adjustbox}{width=\textwidth, totalheight=\dimexpr\textheight-\ht\fullpagecapbox-\dp\fullpagecapbox-8pt\relax, keepaspectratio}
\setlength{\tabcolsep}{3.2pt}
\begin{tabular}{lccccccccc|cc}
\toprule
Model & Closer & Closest-3 & Wall & Count & Feasible & Left/Right & Near & Occlusion & Depth & Avg & Cat-avg \\
\midrule
\textbf{FlowBench-Sim} \\
\midrule
\textit{Chance} & 25.0 & 5.0 & 50.0 & 25.0 & 50.0 & 50.0 & 50.0 & 50.0 & -- & 35.8 & 38.1 \\
\midrule
Human$^\ddagger$ & 70.6 & 25.0 & 90.9 & 50.0 & 73.3 & 86.7 & 46.7 & 80.0 & -- & 64.2 & 65.4 \\
\midrule
\multicolumn{12}{l}{\textit{Proprietary}} \\
GPT-6 Astra (max)$^\P$ & 75.0 & 23.3 & 77.8 & \textbf{39.7} & \textbf{65.5} & \textbf{98.0} & 61.8 & 73.9 & -- & \textbf{60.2} & \textbf{64.4} \\
GPT-6 Astra (low)$^\P$ & 77.3 & 23.3 & 66.7 & 19.0 & 60.3 & \textbf{98.0} & 60.0 & 56.5 & -- & 54.6 & 57.6 \\
GPT-5.6 Luna (low)$^\S$ & 44.4 & 25.9 & 50.0 & 33.3 & 55.0 & 62.5 & \textbf{63.2} & \textbf{83.3} & -- & 47.6 & 52.2 \\
Claude-Fable-5.1 (low)$^\S$ & \textbf{78.9} & \textbf{33.3} & \textbf{100.0} & 30.8 & 53.8 & 90.5 & 52.2 & 57.1 & -- & 56.6 & 62.1 \\
\midrule
\multicolumn{12}{l}{\textit{Open-source}} \\
Qwen3-VL-8B~\cite{bai2025qwen3} & 47.2 & 11.9 & \textbf{52.6} & 24.5 & 56.6 & 77.3 & 58.4 & 55.0 & -- & 46.9 & 47.9 \\
Qwen3-VL-32B~\cite{bai2025qwen3} & 41.8 & \textbf{17.0} & 38.6 & 20.9 & 67.6 & 81.9 & 55.3 & \textbf{62.5} & -- & 48.3 & 48.2 \\
Qwen3.5-9B~\cite{qwen35} & 53.5 & 10.7 & \textbf{52.6} & 23.3 & 52.8 & 68.4 & 53.3 & 51.7 & -- & 44.3 & 45.8 \\
InternVL3.5-8B~\cite{wang2025internvl3} & 42.9 & 9.5 & 29.8 & 16.1 & 54.7 & 65.5 & 51.0 & 55.0 & -- & 40.6 & 40.6 \\
InternVL3.5-30B-A3B~\cite{wang2025internvl3} & 32.6 & 9.2 & 36.8 & 14.2 & 66.7 & 67.0 & 53.0 & 53.3 & -- & 41.4 & 41.6 \\
LLaVA-OV-1.5-8B~\cite{an2025llavaov} & 47.5 & 5.7 & 29.8 & 20.6 & 71.1 & \textbf{86.2} & 51.9 & 55.0 & -- & 47.2 & 46.0 \\
LLaVA-OV-2-8B~\cite{an2026llavaov} & 48.9 & 14.3 & 35.1 & \textbf{33.6} & \textbf{74.8} & 78.2 & 57.5 & 55.0 & -- & \textbf{51.1} & \textbf{49.7} \\
Gemma-4-12B~\cite{team2026gemma} & 42.6 & 13.1 & 31.6 & 19.4 & 61.6 & 50.9 & \textbf{59.0} & 55.0 & -- & 41.6 & 41.6 \\
Gemma-4-31B~\cite{team2026gemma} & \textbf{58.2} & 13.1 & \textbf{52.6} & 26.4 & 55.7 & 57.5 & 53.6 & 54.2 & -- & 44.6 & 46.4 \\
\midrule
\multicolumn{12}{l}{\textit{Fine-tuned on FlowBench-Sim-Train}} \\
Qwen3-VL-8B FT & 89.0 & 36.3 & 93.0 & \textbf{57.3} & 88.4 & \textbf{98.6} & \textbf{88.0} & 72.5 & -- & 76.3 & 77.9 \\
Qwen3.5-9B FT & \textbf{91.1} & \textbf{40.5} & 93.0 & 56.4 & 85.8 & 96.0 & 86.3 & \textbf{82.5} & -- & 76.6 & \textbf{79.0} \\
LLaVA-OV-2-8B FT & 89.4 & 39.3 & \textbf{94.7} & \textbf{57.3} & \textbf{89.6} & 98.0 & 86.3 & 72.5 & -- & \textbf{76.7} & 78.4 \\
PSI-0.6~\cite{psi} & 85.1 & 33.9 & 73.7 & 48.2 & 85.9 & 97.4 & 87.2 & 77.5 & -- & 73.1 & 73.6 \\
PSI-0.6 + WM$^\dagger$ & 27.7 & 8.9 & 63.2 & 31.8 & 76.4 & 90.5 & 60.7 & 55.0 & -- & 50.7 & 51.8 \\
\midrule[\heavyrulewidth]
\textbf{FlowBench-Real} \\
\midrule
\textit{Chance} & 26.7 & 7.0 & -- & 25.1 & -- & 50.0 & 50.0 & 50.0 & 50.0 & 37.6 & 37.0 \\
\midrule
Human$^\ddagger$ & 87.8 & 87.2 & -- & 63.5 & -- & 86.7 & 81.0 & 73.6 & 70.4 & 77.2 & 78.6 \\
\midrule
\multicolumn{12}{l}{\textit{Proprietary}} \\
GPT-6 Astra (max)$^\P$ & \textbf{93.5} & 95.4 & -- & \textbf{82.3} & -- & \textbf{95.0} & \textbf{88.6} & 86.7 & \textbf{83.3} & \textbf{89.9} & 89.2 \\
GPT-6 Astra (low)$^\P$ & 92.4 & 95.4 & -- & \textbf{82.3} & -- & \textbf{95.0} & 87.6 & 84.4 & 76.4 & 88.4 & 87.6 \\
GPT-5.6 Luna (low)$^\S$ & 77.8 & 63.3 & -- & 69.6 & -- & 61.9 & 69.7 & 50.0 & 50.0 & 65.0 & 63.2 \\
Claude-Fable-5.1 (max)$^\S$ & 90.9 & \textbf{96.3} & -- & 82.1 & -- & 91.1 & 87.2 & \textbf{100.0} & 79.3 & 88.9 & \textbf{89.6} \\
Claude-Fable-5.1 (low)$^\S$ & 87.9 & 85.2 & -- & 75.0 & -- & 93.3 & 84.6 & 81.3 & 82.8 & 85.3 & 84.3 \\\midrule
\multicolumn{12}{l}{\textit{Open-source}} \\
Qwen3-VL-8B~\cite{bai2025qwen3} & 42.3 & 7.9 & -- & 38.2 & -- & 52.2 & 48.9 & 37.2 & \textbf{62.1} & 42.7 & 41.2 \\
Qwen3-VL-32B~\cite{bai2025qwen3} & 40.7 & \textbf{19.0} & -- & \textbf{43.5} & -- & 58.9 & 49.9 & 39.2 & 50.0 & 44.4 & 43.0 \\
Qwen3.5-9B~\cite{qwen35} & 40.6 & 11.4 & -- & 30.0 & -- & 52.2 & 56.1 & 50.7 & 50.0 & 42.7 & 41.6 \\
InternVL3.5-8B~\cite{wang2025internvl3} & 42.9 & 10.1 & -- & 38.9 & -- & 42.7 & 54.7 & 48.6 & 55.9 & 42.7 & 42.0 \\
InternVL3.5-30B-A3B~\cite{wang2025internvl3} & 30.9 & 10.6 & -- & 42.2 & -- & 51.1 & 51.1 & 50.0 & 50.0 & 41.2 & 40.9 \\
LLaVA-OV-1.5-8B~\cite{an2025llavaov} & 36.5 & 9.9 & -- & 38.4 & -- & 45.1 & 53.0 & 48.6 & 51.8 & 41.0 & 40.5 \\
LLaVA-OV-2-8B~\cite{an2026llavaov} & 34.7 & 12.8 & -- & 43.3 & -- & 56.1 & 55.8 & 47.9 & 58.9 & 45.0 & 44.2 \\
Gemma-4-12B~\cite{team2026gemma} & 41.4 & 10.4 & -- & 38.2 & -- & 59.9 & 54.2 & 49.0 & 50.0 & 44.6 & 43.3 \\
Gemma-4-31B~\cite{team2026gemma} & \textbf{59.3} & 16.5 & -- & 21.9 & -- & \textbf{64.1} & \textbf{59.8} & \textbf{70.8} & 52.3 & \textbf{50.4} & \textbf{49.2} \\
\midrule
\multicolumn{12}{l}{\textit{Fine-tuned on FlowBench-Sim-Train}} \\
Qwen3-VL-8B FT & 61.4 & 58.0 & -- & 29.5 & -- & 70.3 & 80.0 & 31.9 & \textbf{62.8} & 59.6 & 56.3 \\
Qwen3.5-9B FT & 69.3 & \textbf{62.0} & -- & 27.5 & -- & \textbf{87.6} & \textbf{84.1} & \textbf{62.2} & 62.6 & \textbf{67.7} & \textbf{65.0} \\
LLaVA-OV-2-8B FT & \textbf{73.0} & 61.0 & -- & \textbf{31.0} & -- & 80.7 & 83.6 & 36.1 & 61.4 & 64.9 & 61.0 \\
PSI-0.6~\cite{psi} & 63.0 & 29.6 & -- & 26.7 & -- & 57.5 & 74.6 & 50.0 & 41.1 & 51.4 & 48.9 \\
\bottomrule
\end{tabular}
\end{adjustbox}
\par\vspace{6pt}\usebox{\fullpagecapbox}
\end{table*}

\subsection{Setup}
\label{sec:exp_setup}

\noindent\textbf{Evaluation modes.}
We compare three modes: \textbf{1A} answers directly from the before image $I_0$, motion $t_m$, and question $t_q$; \textbf{1B} uses the same input with reasoning before answering; and \textbf{2A} answers the corresponding static question from the after image $I_1$, without the motion.
The accuracy difference between 2A and 1A measures the imagination gap.
All evaluations use colored Set-of-Marks (SoM) dots~\cite{yang2023setofmark}.

\noindent\textbf{Models.}
We evaluate three proprietary VLMs (GPT-6 Astra, GPT-5.6 Luna, and Claude-Fable-5.1) and nine open-source VLMs. We also test with PSI-0.6~\cite{psi}, a model that unifies VLM and world modeling abilities (Suppl.~\ref{app:psi}).
Proprietary models use category-stratified question subsets; open-source results average three answer-order shuffles.
For proprietary models, direct answering does not imply disabled internal reasoning.
The full model list and inference protocols appear in Suppl.~\ref{supp:exp_setup}.

\noindent\textbf{Fine-tuning.}
We fine-tune Qwen3-VL-8B, Qwen3.5-9B, and LLaVA-OV-2-8B on FlowBench-Sim-Train for direct answering, and additionally train a Qwen3-VL-8B reasoning variant using $c_0$.
No real data is used, so FlowBench-Real-Test measures sim-to-real transfer.
Training details appear in Suppl.~\ref{supp:fine-tune}.

\noindent\textbf{Human baseline.}
Twelve participants provide 403 responses to 240 FlowBench-Real-Test questions and 120 responses on FlowBench-Sim-Test, using the same inputs as 1A. We kindly refer readers to Suppl.~\ref{supp:exp_setup} for
the protocol and agreement statistics.

\subsection{Results}
\label{sec:exp_results}
\cref{tab:results_1a} reports 1A accuracy for all models. We emphasize FlowBench-Real-Test as the benchmark for real-world capability and report FlowBench-Sim-Test as an indicator for simulated scenes. \cref{tab:results_settings} compares the three modes for Qwen3-VL-8B before and after fine-tuning. Further results analysis is in Suppl.~\ref{app:extra_results}.

\noindent\textbf{Main results.}
Base open-source models are close to chance on both benchmarks (40.6--51.1\% on FlowBench-Sim and 41.0--50.4\% on FlowBench-Real, against 35.8\% and 37.6\%), and scale helps little. Proprietary models are much stronger on FlowBench-Real, where GPT-6 Astra and Claude-Fable-5.1 reach 88.4\% and 85.3\% at low effort, but remain weak on FlowBench-Sim (54.6\% and 56.6\%), where Closest Three and Count Near stay near or below chance. The smaller GPT-5.6 Luna is much weaker on both (65.0\% and 47.6\%).

\noindent\textbf{Categories.}
Closest Three is the hardest category for open-source models, which score at most 17.0\% on Sim and 19.0\% on Real (chance 5.0\% and 7.0\%), and Left Right the easiest, with up to 86.2\% and 64.1\%. On Sim Count Near, seven of the nine score below chance, and on the binary Real categories most stay near 50\%, with four scoring exactly 50.0\% on Depth Order.

\noindent\textbf{Imagining versus observing.}
On FlowBench-Real, most models are less accurate when they must imagine the outcome than when they see it (\cref{fig:teaser}): by 5.0 points for Claude-Fable-5.1 at low effort, 3.1 for GPT-6 Astra even at max effort, and 13.0 for Qwen3-VL-8B. GPT-5.6 Luna is the exception and is less accurate when shown the outcome (60.4\% vs.\ 65.0\%). On FlowBench-Sim, seeing the outcome does not help and even lowers the accuracy of all proprietary models, whose difficulty there lies in reading the scene (\cref{sec:analysis_sim}). On Real, CSI thus emerges as a missing component of frontier models' increasingly strong visual abilities (\cref{sec:analysis_bottleneck}).

\noindent\textbf{Models versus humans.}
Humans reach 77.2\% on FlowBench-Real (95\% Wilson interval 72.8--81.0\%), 26.8 points above the best open-source model. GPT-6 Astra and Claude-Fable-5.1 exceed them even at low effort, by 11.2 and 8.1 points, whereas GPT-5.6 Luna falls 12.2 points below. On FlowBench-Sim, humans (64.2\%, 55.3--72.2\%) lead every base model, although the fine-tuned models exceed them by more than 12 points (\cref{sec:analysis_human}).

\noindent\textbf{Fine-tuning on FlowBench-Sim-Train.}
\label{sec:exp_training}
Fine-tuning on FlowBench-Sim-Train raises the three models to 76.3--76.7\% on FlowBench-Sim, 25.6 to 32.3 points above their base models and more than 16 points above GPT-6 Astra at max effort. The gains transfer to FlowBench-Real without real training data, by 16.9 to 25.0 points (up to 67.7\% for Qwen3.5-9B), most strongly on Closest Three (from at most 12.8\% to 58.0--62.0\%) and Near Object (+27.8 to +31.1). This growth is achieved without any training on Real dataset, which is surprising for a large domain gap in our settings, as explained in \cref{sec:sim-to-real}. The fine-tuned models still trail humans and frontier models on Real. PSI-0.6 reaches 73.1\% on Sim but only 51.4\% on Real. Training thus narrows the gap for open-source VLMs, and FlowBench-Sim-Train provides scalable training data for CSI with strong sim-to-real transfer.

\begin{table}[t]
\centering
\small
\setlength{\tabcolsep}{2.0pt}
\begin{tabular}{llccc|cc|cc}
\toprule
\multirow{2}{*}{Model} &
\multirow{2}{*}{Set.} &
\multirow{2}{*}{Img.} &
\multirow{2}{*}{Act.} &
\multirow{2}{*}{CoT} &
\multicolumn{2}{c|}{FlowBench-Sim} &
\multicolumn{2}{c}{FlowBench-Real} \\
 & & & & & Avg & Cat. & Avg & Cat. \\
\midrule
Base & 1A & $I_0$ & \checkmark & -- & 46.9 & 47.9 & 42.7 & 41.2 \\
 & 2A & $I_1$ & -- & -- & 50.4 & 52.3 & 55.7 & 53.0 \\
 & 1B & $I_0$ & \checkmark & \checkmark & 43.4 & 43.1 & 49.8 & 48.4 \\
\midrule
FT & 1A & $I_0$ & \checkmark & -- & \textbf{76.3} & \textbf{77.9} & 59.6 & 56.3 \\
 & 2A & $I_1$ & -- & -- & 73.9 & 76.7 & \textbf{73.6} & \textbf{70.3} \\
 & 1B & $I_0$ & \checkmark & \checkmark & 57.4 & 58.5 & 47.8 & 47.0 \\
\midrule
\multicolumn{4}{l}{\textit{$\Delta$ Base 2A$-$1A}} && +3.5 & +4.3 & +13.1 & +11.7 \\
\multicolumn{4}{l}{\textit{$\Delta$ Base 1B$-$1A}} && \textcolor{black}{$-$3.5} & \textcolor{black}{$-$4.9} & +7.1 & +7.2 \\
\multicolumn{4}{l}{\textit{$\Delta$ FT$-$Base, 1A}} && +29.5 & +29.9 & +16.9 & +15.0 \\
\bottomrule
\end{tabular}
\caption{\textbf{Ablation experiments.} Qwen3-VL-8B-Instruct, base and fine-tuned on FlowBench-Sim-Train (FT), in the three settings.
Avg is per question and Cat.\ the mean of the category accuracies; mean over three answer-order shuffles. }
\label{tab:results_settings}
\end{table}

\subsection{Ablation Study}
\label{sec:exp_ablation}
To further investigate what fine-tuning on FlowBench-Sim-Train changes, we cross-train (base vs.\ FT) with the three evaluation modes for Qwen3-VL-8B (\cref{tab:results_settings}).

\noindent\textbf{Fine-tuning improves perception and imagination.}
Fine-tuning raises accuracy both when the model imagines the outcome (1A: +29.5 on Sim, +16.9 on Real) and when it sees it (2A: +23.5 and +17.9). On Sim, it is even slightly more beneficial imagining than seeing (76.3\% vs.\ 73.9\%), closing the imagination gap in the training domain. This can be attributed to biases in training.
On Real, the gap persists (14.0 points, vs.\ 13.1 before): sim-to-real transfer mainly improves reading spatial relations in real images, while imagining motions in them remains hard.

\noindent\textbf{Reasoning traces are a weaker training signal.}
The model trained on reasoning annotations reaches 57.4\% on Sim and 47.8\% on Real in 1B, 18.9 and 11.8 points below the direct-answer model. On Real, it is even 2.0 points below the base model prompted to reason (49.8\%), so textual reasoning learned in simulation does not transfer to real images. This casts doubt on text CoT's ability to solve some vision-centric questions like CSI, suggesting alternative reasoning method, e.g., visual reasoning.

\subsection{World-Model Training on FlowBench}
\label{sec:exp_wm_train}

\begin{table}[t]
\centering
\small
\setlength{\tabcolsep}{2pt}
\resizebox{\columnwidth}{!}{%
\begin{tabular}{l cccccc}
\toprule
 & Acc & \multicolumn{2}{c}{rEPE$_{\text{move}}$} & Blank & Dir & $\Delta$PSNR \\
\cmidrule(lr){3-4}
Method & (\%)$\uparrow$ & mean$\downarrow$ & med$\downarrow$ & (\%)$\downarrow$ & (\textdegree)$\downarrow$ & $\uparrow$ \\
\midrule
Zero flow & 0.0 & 1.000 & 1.000 & 100.0 & -- & 0.00 \\
\midrule
Modular~\cite{liu_grounding_2024,ravi_sam_2025,Bochkovskii2024:arxiv} & 50.2 & 0.581 & 0.398 & 14.8 & \textbf{1.4} & -- \\
\quad + depth calibration & 50.8 \gain{+0.6} & 0.552 & \underline{0.358} & 15.0 & \textbf{1.4} & -- \\
\midrule
FOFPred~\cite{ran_fofpred_2026} & 0.2 & 0.985 & 1.000 & \textbf{0.0} & 39.5 & -- \\
\quad + LoRA~\cite{hu2021lora} & 2.4 \gain{+2.3} & 0.915 & 0.958 & \textbf{0.0} & 25.9 & -- \\
\midrule
Qwen-Image-Edit~\cite{wu_qwenimage_2025} & 2.9 & 2.432 & 1.246 & 2.7 & 82.4 & -1.81 \\
\quad + QLoRA~\cite{dettmers2023qlora} & 7.9 \gain{+5.0} & 2.154 & 1.014 & 2.9 & 64.8 & -1.63 \\
\midrule
InstructPix2Pix~\cite{brooks2023instructpix2pix} & 0.5 & 1.620 & 1.000 & 66.7 & 82.1 & -4.92 \\
\quad + plain MSE & 10.5 \gain{+10.0} & 2.659 & 1.221 & 2.9 & 71.0 & -3.34 \\
\quad + SoM target & 13.7 \gain{+13.2} & 2.938 & 1.122 & 3.1 & 64.4 & -0.97 \\
\quad + mask-weighted & 16.4 \gain{+15.9} & 2.655 & 1.000 & 2.7 & 57.3 & -3.99 \\
\midrule
$\Psi$-0.6~\cite{psi}, default seq 32 & 10.8 & 0.858 & 0.997 & \textbf{0.0} & 51.4 & +0.79 \\
\quad + FlowBench-Sim & 36.0 \gain{+25.2} & 0.603 & 0.600 & 27.5 & 15.1 & \textbf{+1.60} \\
\midrule
$\Psi$-0.6 + pMotion + CFG2.0~\cite{ho2022cfg} & \underline{52.1} & \underline{0.534} & 0.372 & \textbf{0.0} & 25.1 & \underline{+1.03} \\
\quad + FlowBench-Sim & \textbf{61.7} \gain{+9.6} & \textbf{0.423} & \textbf{0.309} & \underline{0.3} & \underline{8.6} & +0.94 \\
\bottomrule
\end{tabular}}
\caption{\textbf{Training world models on FlowBench-Sim.} Text-to-flow and text-conditioned frame generation on the FlowBench-Sim test split, zero-shot and after training on FlowBench-Sim-Train (ProcTHOR). Acc = percentage of samples with per-pixel rEPE$_{\text{move}}\le 0.40$; rEPE$_{\text{move}}$ = per-pixel relative endpoint error on ground-truth moving pixels (mean and median over samples); Blank = percentage of samples with no predicted motion ($<$1\,px); Dir = 2-D angle between predicted and ground-truth motion where both move; $\Delta$PSNR = PSNR of the generated frame on the change region minus that of copying frame 0, in each model's own generation domain. The modular pipeline receives camera intrinsics, and the image editors the ground-truth frame-0 mask.}
\label{tab:flow_epe_expert_sim_training}
\end{table}

Beyond CSI questions, FlowBench-Sim pairs every text motion command $t_m$ with simulator-exact supervision of its outcome: the before and after frames $(I_0,I_1)$ and the optical flow $F_{01}$ computed from object metadata and depth (\cref{sec:dataset_construction}). These are the inputs and targets of two world-model abilities, predicting the flow of a described motion (text-to-flow) and generating the frame after it (text-conditioned frame generation). We therefore ask whether FlowBench-Sim-Train is also a useful training corpus for world models.

\noindent\textbf{Setup.}
We train four families of models that predict motion from text and an image: the flow predictor FOFPred~\cite{ran_fofpred_2026} with LoRA~\cite{hu2021lora}, the image editors Qwen-Image-Edit~\cite{wu_qwenimage_2025} with QLoRA~\cite{dettmers2023qlora} and InstructPix2Pix~\cite{brooks2023instructpix2pix} with three training objectives, and the unified world model $\Psi$-0.6~\cite{psi} with its default decoding and with motion-first decoding and classifier-free guidance~\cite{ho2022cfg}. Each model is trained on ProcTHOR scenes from FlowBench-Sim-Train and evaluated on the held-out iTHOR scenes of FlowBench-Sim-Test, so its gains must generalize to a new scene source. We compare against a zero-flow baseline and a modular pipeline of Grounding DINO~\cite{liu_grounding_2024}, SAM~2~\cite{ravi_sam_2025}, and Depth Pro~\cite{Bochkovskii2024:arxiv} that receives camera intrinsics as oracle input. 

\noindent\textbf{Training improves every model family.}
As shown in \cref{tab:flow_epe_expert_sim_training}, training raises text-to-flow accuracy for all four model families: FOFPred from 0.2\% to 2.4\%, Qwen-Image-Edit from 2.9\% to 7.9\%, InstructPix2Pix from 0.5\% to up to 16.4\%, and $\Psi$-0.6 from 10.8\% to 36.0\% with default decoding and from 52.1\% to 61.7\% with motion-first decoding and guidance. The predicted motion direction also becomes more accurate: the angular error of $\Psi$-0.6 falls from 51.4\textdegree{} to 15.1\textdegree{} and from 25.1\textdegree{} to 8.6\textdegree{}, and that of FOFPred from 39.5\textdegree{} to 25.9\textdegree{}. The trained $\Psi$-0.6 reaches the highest accuracy and the lowest endpoint error in the table (61.7\%, mean rEPE$_{\text{move}}$ 0.423), surpassing the modular pipeline (50.8\%) despite its oracle input.

\noindent\textbf{Frame generation.}
For the generated outcome frame, training improves $\Delta$PSNR on the change region for Qwen-Image-Edit ($-1.81$ to $-1.63$), InstructPix2Pix ($-4.92$ to between $-3.99$ and $-0.97$), and $\Psi$-0.6 with default decoding (+0.79 to +1.60), while $\Psi$-0.6 with motion-first decoding stays roughly unchanged (+1.03 to +0.94). Only $\Psi$-0.6 generates frames closer to the true outcome than a copy of the before frame ($\Delta$PSNR $>0$).

\noindent\textbf{FlowBench as a training corpus for world models.}
These gains rest on FlowBench's alignment: each text command is paired with the exact flow and outcome frame of the motion it describes, whereas video corpora require estimated flow and motion descriptions written after the fact, and the planner covers diverse objects, directions, and distances across 10K scenes. Because all four model families improve with the same data and the gains transfer from ProcTHOR to iTHOR, FlowBench-Sim-Train provides training signal for text-conditioned world modeling beyond the CSI questions themselves. Better world-modeling abilities are in turn what world-model-assisted answering needs (\cref{sec:analysis_wm}).

\section{Analysis}

\subsection{Is CSI the bottleneck of the task?}
\label{sec:analysis_bottleneck}

Modes 1A and 2A differ only in whether the outcome must be imagined from $(I_0,t_m)$ or is seen in $I_1$, so their gap measures the cost of imagination. On FlowBench-Real, imagining costs GPT-6 Astra 3.1 points at max effort, Claude-Fable-5.1 5.0 points at low effort, and Qwen3-VL-8B 13.0 points, a gap that fine-tuning on simulation leaves unchanged (14.0; \cref{fig:teaser,tab:results_settings}). For the two strongest proprietary models, this is about 30\% of their 1A errors. Text reasoning narrows the gap without closing it: at max effort, GPT-6 Astra and Claude-Fable-5.1 remain 3.1 and 1.4 points below their after-image accuracy, and the fine-tuned model trained on reasoning traces remains 25.8 points below. CSI is thus a bottleneck on real scenes that textual reasoning does not remove, which motivates imagining the outcome visually, as we explore with world models below.
Not every model follows this pattern: GPT-5.6 Luna is less accurate when shown the outcome, and on FlowBench-Sim seeing the outcome lowers the accuracy of all proprietary models, whose bottleneck there lies in reading metric relations from the image (\ref{sec:analysis_sim}). We report these cases and per-category results in Suppl.~\ref{app:gpt_modes} (\cref{tab:gpt_modes}). As FlowBench-Sim mainly serves training, we focus on Real for this question.

\subsection{Why is Sim hard for frontier models?}
\label{sec:analysis_sim}
At low effort, GPT-6 Astra and Claude-Fable-5.1 score about 30 points lower on FlowBench-Sim than on FlowBench-Real (54.6\% vs.\ 88.4\% and 56.6\% vs.\ 85.3\%), whereas open-source models, near chance on both, differ by at most 6.2 points (\cref{tab:results_1a}). The drop concentrates on categories that depend on metric distances between objects, such as Closest Three and Count Near, while Left Right stays above 90\%. Seeing the outcome does not help significantly on these categories either, so the bottleneck is estimating distances in a single rendering rather than imagining the motion. We attribute this mainly to the relatively smaller moving regions and more delicate metric distances. Humans, who cannot have seen our data, also lose 13 points on Sim, mostly on the same categories, which supports this explanation over \emph{exposure} of frontier models to the public robot videos behind Real, although exposure may add to their larger drop. Per-category analysis is in Suppl.~\ref{app:sim_hard}.

\begin{figure}[t]
  \centering
  \includegraphics[width=1.0\columnwidth]{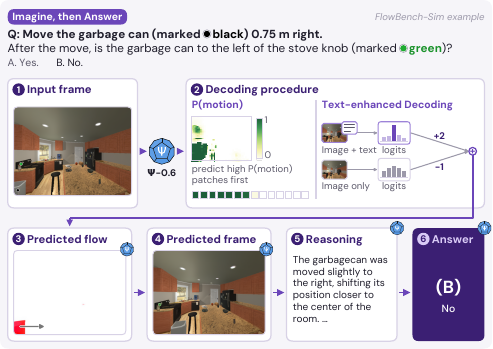}
\caption{\textbf{World-model-assisted VQA.} $\Psi$-0.6~\cite{psi} first imagines the motion as optical flow and a future
frame, decoding high-$P(\mathrm{motion})$ patches first with text-enhanced guidance ($+2/-1$), then reasons
over the imagined outcome to answer. One model performs every step ($\Psi$ badges).}
  \label{fig:wm_vlm}
\end{figure}

\subsection{Strong Sim-to-Real Transfer with FlowBench}
\label{sec:sim-to-real}
One of the surprising findings of this paper is the strong native sim-to-real demonstrated by the FlowBench-Sim-Train dataset. There is a significant domain gap in both the visual style (AI2-THOR simulated and real-world recordings), visual contents (room-level navigation and tabletop manipulation), text questions (short and long) and even question categories as shown in Figs.~\ref{fig:vis_sim_1}-\ref{fig:vis_real_4}. However, all the VLMs that are fine-tuned on FlowBench-Sim-Train directly demonstrate 16.9 to 25.0 percentage points of growth on FlowBench-Real-Test. We therefore attribute this growth to a real ability growth instead of merely task alignment. This result is significant in 1) the ``imagination gap" inside VLMs is at least partially due to lack of specified training, and our proposed FlowBench is helpful for the training recipe; 2) the automatic data collection pipeline we proposed produces high-quality data, which can be claimed as a general data desgin method for other tasks.

\subsection{When does WM reasoning help or hurt?}
\label{sec:analysis_wm}
Since text reasoning does not close the imagination gap, we ask whether imagining the outcome \emph{visually} helps. PSI-0.6~\cite{psi} models language and vision with next-token prediction and can generate visual intermediates; we call its text-to-flow and flow-to-RGB abilities world-model (WM) abilities. In imagine-then-answer mode (\cref{fig:wm_vlm}), it predicts the flow $\hat{F}_{01}$ of the motion, generates the outcome frame $\hat{I}_1$, and answers from it: $(I_0,t_m,t_q)\to\hat{F}_{01}\to\hat{I}_1\to t_a$. During decoding, it generates high-motion patches first and strengthens the motion description with text-enhanced guidance. Each answering mode uses its own training recipe (details in Suppl.~\ref{app:wm}).

\noindent\textbf{Imagining does not yet help.}
On FlowBench-Sim, this mode reaches 50.7\%, below PSI-0.6 trained for direct answering (73.1\%) and below the same model answering directly (55.7\%). The drop concentrates on Count Near ($-18.2$) and Closer To ($-10.6$), whereas Contact Wall, Feasibility, and Left Right are unchanged.

\noindent\textbf{Diagnosis.}
The pattern points to the quality of the imagined intermediates rather than to visual imagination itself. The categories that degrade most require every marked object to keep its identity and precise position in $\hat{I}_1$, as Count Near compares the mover with six listed objects, whereas coarse relations such as Left Right survive imperfect renderings. Generated frames often violate this requirement: in \cref{fig:wm_vlm}, the dot marking the garbage can is missing from the predicted frame, so the question can no longer be grounded in it. Since seeing the true outcome helps strong models (\cref{sec:analysis_bottleneck}), we expect world models that preserve such details to make visual imagination useful; current WM-assisted VLMs fall short of this. FlowBench-Sim-Train can help close this gap: training on it raises the text-to-flow accuracy of $\Psi$-0.6 on held-out scenes from 52.1\% to 61.7\%, with gains for every model family we test (\cref{sec:exp_wm_train}).

\subsection{Human Performance Analysis}
\label{sec:analysis_human}
On FlowBench-Real, GPT-6 Astra is more accurate than pooled human accuracy in every category even at low effort, an interesting finding complementary to existing spatial benchmarks~\cite{yang2025thinking,yin2025mindcube}; on FlowBench-Sim, humans lead all base models. We examine where humans fail (details in Suppl.~\ref{app:human}); since humans and models answered different subsets of questions, these comparisons are indicative.

\noindent\textbf{Humans struggle with metric thresholds.}
Humans answer relational questions (Closer To, Closest Three, Left Right) correctly 87.2\% of the time but threshold questions (Count Near, Near Object) only 71.3\% of the time, whereas frontier models drop less, from 94.2\% to 85.6\% for GPT-6 Astra and from 89.5\% to 80.6\% for Claude-Fable-5.1 at low effort; the advantage of frontier models over humans is largest on the latter (14.3 points for GPT-6 Astra and 9.3 for Claude-Fable-5.1, against 7.0 and 2.3 on relations). On FlowBench-Sim, humans lose most on the same distance-based categories, from 87.2\% to 25.0\% on Closest Three and from 81.0\% to 46.7\% on Near Object, whereas Left Right stays at 86.7\%. Two participants also reported that judging centimeter thresholds from a single image felt unnatural.
\noindent\textbf{Humans answer as if nothing moved.}
Humans are correct on 84\% of questions whose answer the motion leaves unchanged but on 68.5\% of those it changes; on changed Depth Order questions they score 54\%, and every error is the before-image answer. Long motions (over 15\,cm) are also harder than short ones (72\% vs.\ 82\%), so human errors come mainly from failing to update the scene, \ie from imagination itself. GPT-6 Astra shows a weaker form of this tendency (Suppl.~\ref{app:change}). Human errors are not label errors: a manual re-check confirmed the label of every question that both participants missed, and on 91\% of questions with two answers at least one participant was correct.

\noindent\textbf{Implications.}
On FlowBench-Real, humans and frontier models have different strengths: the models are better at the metric judgments that humans find hard. The models reach this without explicit 3D inputs or geometric modules, which complements approaches that add such inductive biases~\cite{hong20233d,chen2024spatialvlm,cheng2024spatialrgpt}; our evaluation cannot tell whether scale, data, or other factors are responsible. Human accuracy is thus a reference point for CSI rather than an upper bound.

\section{Conclusion}

We formalized Counterfactual Spatial Imagination (CSI), the ability to infer how the spatial relations in a scene change under a hypothetical object motion, and introduced FlowBench, a benchmark and training resource spanning simulated homes and real robot tabletops. Its pipelines are automatic, with verified test answers, and the simulated pipeline scales to 420K training questions with outcome images, flow, depth, segmentation, and reasoning annotations, so the same data supervises both VLMs and world models. For VLMs, our experiments lead to three findings. First, frontier models retain an imagination gap that textual reasoning narrows but does not close, even on FlowBench-Real, where they exceed pooled human accuracy; on FlowBench-Sim, humans remain ahead of all base models. Second, humans and models fail differently: humans tend to answer biased to the input state, whereas the models are better at metric judgments. Third, open-source models remain far below humans, but fine-tuning on FlowBench-Sim-Train improves them by up to 32.3 points on Sim and 25.0 on Real, a strong sim-to-real transfer. For world models, training on FlowBench-Sim improves text-to-flow prediction for every model family we test, by up to 25.2 points, yet world-model-assisted answering does not yet help, as imagined intermediates lose key details. FlowBench thus serves both as a training corpus for world models and as a test of when their imagination helps VLM reasoning. We hope FlowBench supports models that imagine spatial change, not only perceive it.

\noindent\textbf{Limitations.}
1A accuracy combines perception and imagination; the 2A control separates them mainly on FlowBench-Real, where static questions are easier, whereas on FlowBench-Sim static metric questions are themselves hard, so we draw conclusions about imagination from the 2A--1A gap on Real. FlowBench-Real answers rely on estimated depth, masks, and tracks, so absolute distances are approximate. For cost reasons, proprietary models answer 20--50\% of the questions, on samples that differ between models. The Real videos are public and may have been seen during training of the tested VLMs. Finally, motions are single rigid translations of one object, without rotations, deformations, or multi-step interactions, and our world-model-assisted answering covers one model and one inference procedure. We discuss these limitations further in Suppl.~\ref{app:limitations}.

\section{Statements}

\noindent\textbf{AI Use Statement}
We used two LLM tools in this work.
Claude Opus 5.5 (Anthropic) was used to help in writing code, running experiments, making figures, providing content of some sections in Supplementary, and polishing the writing of the main paper.
GPT-6 Astra (OpenAI) was used to polish the writing of the paper.
All code, experimental results, figures, and text produced or edited with these tools were reviewed and verified by the authors, who take full responsibility for the content of this paper.
Separately, the models we evaluate on FlowBench, including GPT-6 Astra, are the subjects of our experiments and are described in \cref{sec:exp_setup}.

\noindent\textbf{Ethics Statement}
To establish a human performance baseline, volunteers from our research lab answered CSI questions through a web application in sessions lasting approximately 20--30 minutes. Participation was voluntary, and all participants provided consent.

\noindent\textbf{Reproducibility Statement}
The code and complete dataset are publicly available at \url{https://github.com/liuyifan22/FlowBench}.

\noindent\textbf{Acknowledgements}
We thank the volunteers from our research lab for their time and participation in the human baseline evaluation.

\noindent\textbf{Funding}
This work was supported by the following awards. D.L.K.Y.: Simons Foundation grant 543061, National Science Foundation CAREER grant 1844724, National Science Foundation Grant NCS-FR 2123963, Office of Naval Research grant S5122, ONR MURI 00010802, ONR MURI S5847, and ONR MURI 1141386 - 493027.
We also thank Stanford HAI, Stanford Data Sciences, Stanford Marlowe team, and the
Google TPU Research Cloud team for computing support.
{
    \small
    \bibliographystyle{ieeenat_fullname}
    \bibliography{main}
}

\clearpage
\setcounter{page}{1}
\renewcommand{\thepage}{S\arabic{page}}
\maketitlesupplementary
\appendix

\section{FlowBench-Sim Construction}

\subsection{Camera Selection Algorithm}
\label{app:camera_selection}

For each scene, we automatically select a small set of informative camera poses. The procedure has four stages: candidate positioning, view scoring, per-room proposal generation, and global selection.

\noindent\textbf{Candidate positioning.}
A good initial view should contain several actionable objects and should not be dominated by a wall. We first query the simulator navigation mesh for all reachable standing locations, $\mathcal{P}=\mathrm{GetReachablePositions}(\mathrm{scene})$, and assign them to rooms $r\in\mathcal{R}$ by floor polygon:
\begin{equation}
\mathcal{P}_r
=
\left\{
p\in\mathcal{P}
\;\middle|\;
(p_x,p_z)\in\mathrm{FloorPolygon}(r)
\right\}.
\end{equation}
Within each room, farthest-point sampling (FPS) selects a spatially diverse subset of standing locations $\{p^{(1)},\ldots,p^{(K)}\}\subseteq\mathcal{P}_r$ with $K\approx10$.
Each location is paired with a discrete camera yaw
\begin{equation}
\psi\in\{0^\circ,45^\circ,\ldots,315^\circ\}
\end{equation}
and a horizon $h$ sampled uniformly from $\{0^\circ,30^\circ\}$.
For each resulting pose $c=(p,\psi,h)$, we teleport the camera to the viewpoint and keep the pose only if it observes at least one movable object at a mid-range distance
\begin{equation}
d_{\min}<d<d_{\max},
\end{equation}
with $(d_{\min},d_{\max})=(1.5\,\mathrm{m},\,4.5\,\mathrm{m})$.

\noindent\textbf{View scoring.}
Let $\mathcal{B}(c)$ be the set of visible movable objects within this distance band, and let $\mathcal{B}_{\mathrm{pref}}(c)\subseteq\mathcal{B}(c)$ be the subset of preferred object types.
Let $\mathcal{V}(c)$ be the set of all visible non-floor objects, and let $\delta(c)$ be the depth of a raycast through the image center ($\mathrm{GetCoordinateFromRaycast}$).
Each candidate receives the score
\begin{equation}
\begin{split}
s(c)
=
4\,|\mathcal{B}_{\mathrm{pref}}(c)|
&+
1.5\,|\mathcal{B}(c)|\\
&+
0.1\,|\mathcal{V}(c)|
+
\phi\!\left(\delta(c)\right),
\end{split}
\end{equation}
where
\begin{equation}
\phi(\delta)
=
\begin{cases}
-5, & \delta=\varnothing,\\
-8, & \delta<0.6\,\mathrm{m},\\
\min(\delta,4), & \text{otherwise}.
\end{cases}
\end{equation}
The depth term thus favors open views and strongly penalizes views whose central ray hits a nearby wall.

\noindent\textbf{Per-room proposals.}
For each room, we keep up to $m\in\{2,3\}$ high-scoring poses,
\begin{equation}
\mathcal{C}_r
=
\left\{
c_r^{(1)},\ldots,c_r^{(m)}
\right\},
\end{equation}
greedily rejecting spatially redundant candidates: a candidate $c$ is rejected if
\begin{equation}
\left\lVert x(c)-x(c')\right\rVert_2 < \tau
\end{equation}
for some previously kept pose $c'$, where $x(c)$ is the 3D camera position and $\tau\approx1\,\mathrm{m}$.
The apartment-level proposal pool is $\mathcal{C}=\bigcup_{r\in\mathcal{R}}\mathcal{C}_r$.

\noindent\textbf{Global selection.}
We select the final set of poses $\mathcal{S}^{\star}\subseteq\mathcal{C}$ under the budget
\begin{equation}
y
=
\min\!\left(10,\,3|\mathcal{R}_{\mathrm{floor}}|\right),
\end{equation}
so $y=10$ by default. Let
\begin{equation}
\mathrm{cov}(\mathcal{S})
=
\left\{
r\in\mathcal{R}
\;\middle|\;
\exists\,c\in\mathcal{S},\,r(c)=r
\right\}
\end{equation}
be the set of rooms covered by the currently selected poses $\mathcal{S}$, and let $\mathcal{U}(\mathcal{S})=\bigcup_{r\notin\mathrm{cov}(\mathcal{S})}\mathcal{C}_r$ be the candidates from uncovered rooms.
We first enforce room coverage by repeatedly adding the highest-scoring candidate from an uncovered room:
\begin{equation}
\begin{aligned}
c^\star
={}&
\arg\max_{c\in\mathcal{U}(\mathcal{S})}\;s(c)\\
\text{s.t.}\;{}&
\left\lVert x(c)-x(c')\right\rVert_2\geq\tau
\quad
\forall c'\in\mathcal{S}.
\end{aligned}
\end{equation}
Once every room is covered or the budget is exhausted, we fill any remaining slots by ranking candidates according to
\begin{equation}
\begin{split}
\kappa(c)
=
s(c)
&+
\lambda\,
\mathbf{1}\!\left[r(c)\notin\mathrm{cov}(\mathcal{S})\right]\\
&+
\frac{0.05\lambda}
{1+n_{r(c)}(\mathcal{S})}
-
\rho_{\mathrm{yaw}}(c;\mathcal{S}),
\end{split}
\end{equation}
where $\lambda\approx10^3$ controls the preference for room diversity, $n_r(\mathcal{S})$ is the number of poses already selected from room $r$, and $\rho_{\mathrm{yaw}}(c;\mathcal{S})$ penalizes same-room views with similar yaw ($|\Delta\psi|<35^\circ$).
If the separation constraint prevents filling the budget, we progressively relax $\tau$ from about $1\,\mathrm{m}$ down to $0.35\,\mathrm{m}$.
Finally, the scene is rendered from every pose $c\in\mathcal{S}^{\star}$, and these observations are passed on to question planning.

\subsection{Planning Algorithms}
\label{app:planning_algorithms}

Random object translations almost never produce the rare geometric outcomes we care about, such as an occlusion along the line of sight, a change of nearest landmark, or contact with a wall.
We therefore \emph{plan} each question: we design a desired after-state, convert it into a teleport in the camera frame, and validate the outcome in the simulator.
If the teleport collides, or the measured answer does not match the class the running balance asks for, Planning retries.
The ground-truth (GT) label is always read from simulator metrics after a successful edit, never from the planner's intent.

\noindent\textbf{Shared camera-frame translations.}
Let $r$ and $f$ be the camera's right and forward unit vectors in the horizontal plane.
A cardinal slide of length $m$ in direction $d\in\{\mathrm{left},\mathrm{right},\mathrm{forward},\mathrm{back}\}$ induces the world displacement
\begin{equation}
\Delta(d,m)
=
m\cdot
\begin{cases}
-r, & d=\mathrm{left},\\
+r, & d=\mathrm{right},\\
+f, & d=\mathrm{forward},\\
-f, & d=\mathrm{back}.
\end{cases}
\end{equation}
Given an object position $p=(p_x,p_y,p_z)$, the planned target keeps the height fixed:
\begin{equation}
p'
=
\bigl(p_x+\Delta_x,\, p_y,\, p_z+\Delta_z\bigr).
\end{equation}
Unless a category overrides this rule, the mover must lie in the same mid-range band as in \cref{app:camera_selection}, $d_{\min}<d_{\mathrm{cam}}(o)<d_{\max}$, and must appear large enough in the image, $m_{\mathrm{ext}}/d_{\mathrm{cam}}>\alpha_{\min}$ with $\alpha_{\min}\approx0.08$.
Generic proposals enumerate the four directions above and lengths $m\in\{0.6,0.5,0.4,0.25\}\,\mathrm{m}$.
Each move is executed as a collision-checked teleport ($\mathrm{force}=\mathrm{false}$); if it is blocked, we binary-search along the motion ray for a few halvings (typically $4$), never raising $p_y$.
For selected categories, pickupable objects may settle under gravity after the teleport.
The mover must remain visible from the fixed camera after the edit.

\noindent\textbf{Class balance.}
For yes/no categories, we maintain running counts $(n_{\mathrm{yes}},n_{\mathrm{no}})$ and a target no-rate $\rho_{\mathrm{no}}$.
The planner prefers the under-represented class; on a tie, it samples
\begin{equation}
\mathbb{P}(\text{prefer yes})
=
1-\rho_{\mathrm{no}}.
\end{equation}
The category-specific targets are
$\rho_{\mathrm{no}}^{\mathrm{occ}}\approx0.20$,
$\rho_{\mathrm{no}}^{\mathrm{wall}}\approx0.08$,
$\rho_{\mathrm{no}}^{\mathrm{feas}}\approx0.55$,
and $\rho_{\mathrm{no}}^{\mathrm{lr}}=\rho_{\mathrm{no}}^{\mathrm{near}}=0.50$.
Flip-seeking categories (closest-three, count-near, near-object, and left/right) aim to change the answer relative to the before-state with probability
\begin{equation}
\pi_{\mathrm{flip}}\approx0.80.
\end{equation}

\subsubsection{Occlusion}
\label{app:plan_occlusion}

\noindent\textbf{Goal.}
A yes/no question: after the move, does a large object $A$ occlude a small object $B$?
Let $M_A^{(1)}$ be the mask of $A$ in the after image and $M_B^{(0)}$ the mask of $B$ in the before image.
The cover rate is
\begin{equation}
\mathrm{cover}(A\!\to\!B)
=
\frac{\bigl|M_A^{(1)}\cap M_B^{(0)}\bigr|}
{\bigl|M_B^{(0)}\bigr|},
\end{equation}
and the GT answer is yes if and only if
\begin{equation}
\mathrm{cover}(A\!\to\!B)\ge\tau_{\mathrm{occ}},
\qquad
\tau_{\mathrm{occ}}=0.30.
\end{equation}

\noindent\textbf{Pair selection.}
We enumerate pairs of visible large occluders (plants, chairs, pans, \ldots) and small occludees (phones, statues, \ldots) that satisfy
\begin{equation}
\begin{aligned}
\|p_A-p_B\|_2&\le 2.25\,\mathrm{m},\\
\bigl|(p_A)_y-(p_B)_y\bigr|&\le 0.85\,\mathrm{m},
\end{aligned}
\end{equation}
and skip occludees with tiny masks (below $\approx150$~px for natural images, relaxed to $\approx20$~px with Set-of-Marks).

\noindent\textbf{Fast ray plan (ProcTHOR).}
Let $C$ be the camera center and $B$ the occludee center, so the unit ray from the camera to $B$ is $\hat{d}=(B-C)/\|B-C\|$.
We place $A$ on this ray just in front of $B$, at distance
\begin{equation}
t_{\mathrm{front}}
=
\|B-C\|-\mu,
\qquad
\mu\approx0.12\,\mathrm{m}.
\end{equation}
Writing $t_A$ for the current distance of $A$ along the ray, the target is $A^{\star}=C+t^{\star}\hat{d}$ with
\begin{equation}
t^{\star}
=
\begin{cases}
t_A, & \text{$A$ already in front of $B$},\\
t_{\mathrm{front}}, & \text{otherwise},
\end{cases}
\end{equation}
and the height is clamped so that $(A^{\star})_y\le(A)_y$.
For a controlled \emph{no} example, we instead offset $A^{\star}$ sideways in the horizontal plane by $\nu\approx0.22\,\mathrm{m}$, orthogonally to $\hat{d}$.
We teleport once ($\mathrm{force}=\mathrm{false}$), re-render, and accept the edit if the measured answer matches the running yes/no balance; otherwise, we keep the best opposite-answer attempt as a fallback (budget $\approx16$ tries).

\noindent\textbf{Full cover-grid plan.}
Outside the fast path, the metric size of a pixel at depth $d_A$ is
\begin{equation}
\mathrm{mpp}
=
\frac{2\,d_A\tan(\mathrm{FOV}/2)}{W},
\end{equation}
with image width $W=800$ and $\mathrm{FOV}=90^\circ$.
From the mask centroids $(u_A,v_A)$ and $(u_B,v_B)$, we estimate a lateral slide $m_{\mathrm{lat}}=|u_B-u_A|\,\mathrm{mpp}$, or a short backward pull when the horizontal gap is negligible.
We then search a grid of scales $\{0.5,0.75,\ldots,2.25\}$ and backward pulls $\{0,0.08,0.15,0.25,0.40,0.55\}\,\mathrm{m}$, clamp the applied length to $[0.15,1.50]\,\mathrm{m}$, and keep the pose with maximum cover (up to $\approx48$ tries).

\subsubsection{Closer-to}
\label{app:plan_closer_to}

\noindent\textbf{Goal.}
A 4-way multiple-choice question: after object $A$ moves, which of four landmarks is nearest to it?
We strongly prefer edits that \emph{flip} the nearest landmark from $B$ to $C$.

\noindent\textbf{Offline plan enumeration.}
We work in the horizontal $(x,z)$ plane and write $a,b,c$ for the centers of $A,B,C$.
A candidate pair must have a clear before-gap and well-separated landmarks:
\begin{equation}
d(A,B)+\gamma < d(A,C),
\qquad
\|c-b\|_2\ge\sigma_{\mathrm{BC}},
\end{equation}
with $\gamma=0.18\,\mathrm{m}$ and $\sigma_{\mathrm{BC}}=0.30\,\mathrm{m}$.
Where available, distances are surface clearances between axis-aligned bounding boxes (AABBs),
\begin{equation}
d_{\mathrm{AABB}}(o,o')
=
\Bigl(\sum_{i\in\{x,y,z\}} g_i^{2}\Bigr)^{1/2},
\end{equation}
where $g_i=\max\bigl(0,\,|c_i-c'_i|-\tfrac12(s_i+s'_i)\bigr)$ is the per-axis gap, and $c$ and $s$ are the box center and extents.
Let $m=\tfrac12(b+c)$ and $\hat{u}=(c-b)/\|c-b\|$.
For fractions $\phi\in\{0.45,0.55,0.60,0.50,0.65,0.40\}$ past the midpoint, we propose
\begin{equation}
a^{\mathrm{ideal}}
=
m+\phi\,\|c-b\|\,\hat{u},
\end{equation}
and scale the displacement so that
\begin{equation}
m_{\mathrm{move}}
=
\|a^{\mathrm{ideal}}-a\|_2
\in[0.25,\,1.05]\,\mathrm{m},
\end{equation}
keeping the height fixed, $(a')_y=(a)_y$.
Plans are scored by how strongly the gap flips,
\begin{equation}
\begin{split}
\kappa
={}&
\bigl(d(A,C)-d(A,B)\bigr)_{\mathrm{before}}\\
&+
\bigl(d(A',B)-d(A',C)\bigr)_{\mathrm{after}}
-
0.15\,m_{\mathrm{move}},
\end{split}
\end{equation}
plus a small bonus for well-separated distractors.
We add two farther distractors $D_1,D_2$, with at most one shelving unit and one drawer, so the answer options are $\{B,C,D_1,D_2\}$.

\noindent\textbf{Validation.}
We teleport according to the plan and recompute all distances.
The edit is accepted if $C$ is the unique nearest of the four landmarks after the move, with a uniqueness margin of $\max(0.02,\,0.25\gamma)$.
The fast mode considers up to $\approx6$ plans and the full search up to $\approx24$.
The GT is the letter of the unique nearest landmark; an auxiliary flag records whether the nearest landmark flipped.

\subsubsection{Contact wall}
\label{app:plan_contact_wall}

\noindent\textbf{Goal.}
A yes/no question: after the move, is object $A$ in contact with a wall?
We compute signed clearances between the object AABB and the scene AABB along the axes $\{\pm x,\pm z\}$ and take their minimum,
\begin{equation}
\mathrm{clr}(A)
=
\min
\bigl\{
\mathrm{clr}_{-x},\mathrm{clr}_{+x},\mathrm{clr}_{-z},\mathrm{clr}_{+z}
\bigr\}.
\end{equation}
The GT answer is yes if and only if
\begin{equation}
\mathrm{clr}(A)_{\mathrm{after}}
<
\tau_{\mathrm{wall}},
\qquad
\tau_{\mathrm{wall}}=0.22\,\mathrm{m}.
\end{equation}

\noindent\textbf{Mover selection and answer balance.}
We prefer movers that are already fairly close to a wall,
\begin{equation}
\mathrm{clr}(A)_{\mathrm{before}}
<
\begin{cases}
0.8\,\mathrm{m}, & \text{fast path},\\
1.5\,\mathrm{m}, & \text{otherwise},
\end{cases}
\end{equation}
and whose after-mask is large enough ($\ge30$~px).
Because contact is rare under random motion, the balance targets only $\rho_{\mathrm{no}}^{\mathrm{wall}}\approx0.08$ ``no'' answers.
Most attempts therefore slide \emph{toward} the nearest wall axis $a^{\star}=\arg\min\mathrm{clr}$; occasional slides away from it (base probability $\approx0.05$, overridden by the balance) create the scarce negatives.

\noindent\textbf{Slide search.}
Let $\eta\approx0.10\,\mathrm{m}$ be a stopping margin and $\varepsilon_{\mathrm{pen}}\approx0.02\,\mathrm{m}$ a penetration tolerance.
The fast path moves toward the wall in a single step of length
\begin{equation}
m
=
\mathrm{clr}(A)-\eta,
\end{equation}
provided that $\mathrm{clr}(A)>\eta$ and $m\ge0.05\,\mathrm{m}$.
Slides away from the wall use $m=\max(0.35,\,|\mathrm{clr}|+0.25)$.
The full path searches lengths in $\{0.15,0.25,\ldots,2.5\}\,\mathrm{m}$ with multi-step binary refinement along the wall axis.
We reject edits that penetrate the wall ($\mathrm{clr}<-\varepsilon_{\mathrm{pen}}$) or make the after-mask collapse.
When possible, the measured answer must match the balance; if the first pass fails, a second pass tries longer slides (fast budget $\approx12$ tries).

\subsubsection{Left / right of landmark}
\label{app:plan_left_right}

\noindent\textbf{Goal.}
A yes/no question: after the move, is mover $A$ to the left (or right) of landmark $L$ in the image?
Let $u(\cdot)$ be the horizontal image coordinate of an object's mask centroid.
The geometric relation is
\begin{equation}
\mathrm{left}(A,L)
\iff
u(A)<u(L).
\end{equation}
The wording (``left of'' vs.\ ``right of'') is chosen so that the textual answer matches the running yes/no balance, without changing the geometry.

\noindent\textbf{Projected flip plans.}
For directions $\{\mathrm{left},\mathrm{right}\}$ and lengths
\begin{equation}
m\in\{0.90,0.80,\ldots,0.30\}\,\mathrm{m}
\end{equation}
(base lengths $\{0.80,\ldots,0.20\}$ plus $0.10\,\mathrm{m}$), we predict the after-coordinate $u_1$ of $A$ at the target pose, with $u_0$ its before-coordinate.
A flip plan is kept only if $A$ changes side relative to $u(L)$ and
\begin{equation}
\begin{aligned}
|u_1-u_0|&\ge 36\,\mathrm{px},\\
\min\bigl(|u_0-u(L)|,\,|u_1-u(L)|\bigr)&\ge 12\,\mathrm{px}.
\end{aligned}
\end{equation}
Controlled non-flips (target rate $\approx0.20$) require a smaller but still clear horizontal span and margin.
We teleport, recompute the centroid $u$, and prefer pickupable movers and true flips ($\pi_{\mathrm{flip}}\approx0.80$).
Up to $\approx24$ plans are tried; if a strict flip is required and none succeeds, the category is skipped for that camera rather than silently substituted.
The GT compares $u(A)_{\mathrm{after}}$ with $u(L)$.

\subsubsection{Feasibility}
\label{app:plan_feasibility}

\noindent\textbf{Goal.}
A yes/no question asked on the \emph{after} image: would a further, explicitly named translation of the mover be collision-free?
Here the edit from the before to the after image is only a setup move (a generic translation, optionally followed by gravity settling).
The real work is choosing a \emph{probe} $\Delta_{\mathrm{probe}}$ from the after pose whose outcome matches the desired yes/no balance ($\rho_{\mathrm{no}}^{\mathrm{feas}}\approx0.55$).

\noindent\textbf{Probe planning.}
From the after pose, we try camera-frame probes whose lengths are drawn from
\begin{equation}
\begin{aligned}
\mathcal{M}_{\mathrm{yes}}&\approx\{0.4,0.5\}\,\mathrm{m},\\
\mathcal{M}_{\mathrm{no}}&\approx\{1.5,1.25,1.0\}\,\mathrm{m}
\end{aligned}
\end{equation}
on the fast path; the full path also uses $\{1.5,\ldots,0.4\}\,\mathrm{m}$ and optional downward probes of $\{0.6,0.5,0.4\}\,\mathrm{m}$.
With probability $\approx0.85$, the probe direction is orthogonal to the setup move.
A probe is feasible if and only if every waypoint along its path teleports successfully ($\mathrm{force}=\mathrm{false}$) and the signed wall clearance stays above $-\varepsilon_{\mathrm{pen}}$.
Hard negatives come from compound two-axis probes and from closed-form wall collisions, \ie, probes toward the nearest wall with $m\ge\max(0.05,\,\mathrm{clr}-\eta)$.
The fast mode allows up to $\approx3$ probe attempts; on failure, the ``no'' lengths may be extended to $\{2.0,1.75,1.5\}\,\mathrm{m}$.
The GT is exactly whether the named probe succeeds, and the question text states that probe's direction and length.

\subsubsection{Closest three}
\label{app:plan_closest_three}

\noindent\textbf{Goal.}
Among six listed objects, which three are closest to the mover after the move?
We prefer edits that change the top-3 set.

\noindent\textbf{Neighbor ranking and candidate pool.}
After a generic translation of $A$, we rank the other referable objects by 3D centroid distance
\begin{equation}
d_{\mathrm{c}}(A,o)
=
\|p_A-p_o\|_2,
\end{equation}
allowing at most one shelving unit and one drawer to limit look-alike furniture.
Let $T_0$ and $T_1$ be the top-3 sets before and after the move.
The six options are their union, padded with the next-nearest neighbors after the move:
\begin{equation}
\mathcal{P}
=
\mathrm{Pad}_6\!\bigl(T_0\cup T_1\bigr).
\end{equation}
The answer set is $T_1\cap\mathcal{P}$, which always contains three letters after the options are shuffled (\eg, $\mathtt{ABD}$).

\noindent\textbf{Flip gate.}
With probability $\pi_{\mathrm{flip}}\approx0.80$, we require
\begin{equation}
T_0\neq T_1;
\end{equation}
otherwise, non-flips are deliberately accepted.
Up to $\approx40$ mover attempts are made, first in flip mode and then relaxed.
The GT uses only after-move centroid distances, with no simulator query beyond the edit itself.

\subsubsection{Count near}
\label{app:plan_count_near}

\noindent\textbf{Goal.}
How many of six listed objects lie within distance $\tau$ of the mover after the move?
The answer is restricted to $\{2,3,4,5\}$.

\noindent\textbf{Threshold search.}
After a generic move, we consider thresholds
\begin{equation}
\tau\in\{0.35,0.5,0.65,0.8,1.0,1.25\}\,\mathrm{m},
\end{equation}
shuffled per attempt.
Let $\mathcal{P}$ be the six nearest eligible neighbors (with the same type caps as above), and for $t\in\{0,1\}$ (before and after) define
\begin{equation}
n_t(\tau)
=
\#\bigl\{o\in\mathcal{P}:\,d_{\mathrm{c}}(A_t,o)\le\tau\bigr\}.
\end{equation}
We require $n_1(\tau)\in\{2,3,4,5\}$ and, with probability $\pi_{\mathrm{flip}}\approx0.80$, also $n_0(\tau)\neq n_1(\tau)$.
Fallbacks first relax the count-change preference and then, if needed, the allowed answer set.
The GT is the integer $n_1(\tau)$; the choices are a shuffle of $\{2,3,4,5\}$, and $\tau$ is stated in the question.

\subsubsection{Near object}
\label{app:plan_near_object}

\noindent\textbf{Goal.}
A yes/no question: after the move, is mover $A$ within distance $\tau$ of a named landmark $L$?
If $L$ is near $A$ both before and after the move, the before image alone suffices, so we prefer edits that flip proximity.

\noindent\textbf{Near-set construction.}
Using the same $\tau$ grid as count-near, we define for $t\in\{0,1\}$
\begin{equation}
S_t(\tau)
=
\bigl\{o:\,d_{\mathrm{c}}(A_t,o)\le\tau\bigr\}.
\end{equation}
We prefer landmarks in the symmetric difference
\begin{equation}
L\in S_0(\tau)\,\triangle\,S_1(\tau),
\end{equation}
whose proximity flips, and otherwise fall back to $S_0(\tau)\cap S_1(\tau)$.
When possible, we also require the after answer
\begin{equation}
\mathbf{1}\!\bigl[d_{\mathrm{c}}(A_1,L)\le\tau\bigr]
\end{equation}
to match the class preferred by the balance.
The constraints are relaxed in the order
$(\mathrm{flip},\mathrm{yes})\to(\neg\mathrm{flip},\mathrm{yes})\to(\mathrm{flip},\cdot)\to(\neg\mathrm{flip},\cdot)$,
with up to $\approx8$ successful edits per stage (about $64$ proposals in total) and a yes/no balance near $50\%$.
The GT is yes if and only if $d_{\mathrm{c}}(A_1,L)\le\tau$.

\noindent\textbf{From plan to annotation.}
Across all categories, Planning only proposes and filters edits.
Querying keeps the camera fixed, executes the teleport, and re-reads masks, clearances, and distances; Annotating derives the answer letter (or letter set) from these metrics alone.
Optional VLM rephrasing may rewrite the question and choices, but never selects the answer.

\section{FlowBench-Real Construction}
\label{app:real}
\begin{figure*}[t]
  \centering
  \includegraphics[width=1.0\textwidth]{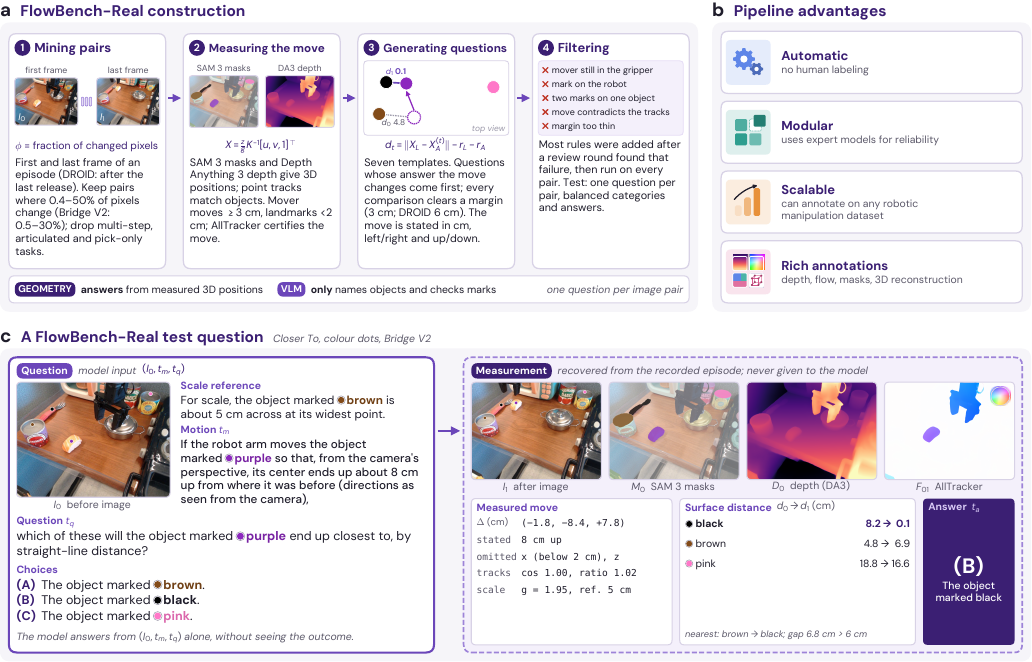}
  \caption{\textbf{FlowBench-Real.}
  \textbf{(a)}~Four-stage curation pipeline; answers come only from the measured 3D geometry, and a VLM only names objects and checks marks.
  \textbf{(b)}~FlowBench-Real test set: 1{,}083 questions, one per image pair, over seven categories and five robot datasets.
  \textbf{(c)}~A test question. The model sees only $(I_0,t_m,t_q)$; the after frame, masks, depth, tracks and distances are what the pipeline measured to derive the answer.}
  \label{fig:flowbench_real}
\end{figure*}

FlowBench-Real replaces Planning and Querying (\cref{app:planning_algorithms}) with measurement: the robot has already executed the move, and we recover it from the recorded video.
Each item is built from an image pair $(I_0,I_1)$ of one episode in four stages: mining the pair (\cref{app:real_mining}), measuring the move (\cref{app:real_measure,app:real_scale}), generating questions from the measured geometry (\cref{app:real_questions}), and filtering and verification (\cref{app:real_filters,app:real_split}).
All quantities are expressed in the camera frame of $I_0$, with $x$ pointing right, $y$ down, and $z$ away from the camera.
The ground-truth answer is always computed from the measured geometry; a VLM only names objects and checks marks, and never selects an answer.

\subsection{Mining image pairs}
\label{app:real_mining}

\noindent\textbf{Frames.}
For every episode, $I_0$ is the first frame.
For Bridge V2, Language-Table, RT-1, and Berkeley UR5, $I_1$ is the last frame.
On Bridge V2 this simple rule accepted more pairs in measurement than an arm-aware frame search ($49$ vs.\ $42$ of $300$ trajectories), because the arm parks and carries objects in the same image region.
For DROID, $I_1$ is never earlier than the last gripper release and is preferably a frame in which the arm has left the view; of the two external cameras, we use the one that sees the release point closer to the image centre.

\noindent\textbf{Episode filters.}
A coarse change filter rejects pairs in which almost nothing or almost everything changes.
With $G_t$ the grayscale frame and $\Omega$ the image interior,
\begin{equation}
\phi
=
\frac{1}{|\Omega|}\sum_{p\in\Omega}\mathbf{1}\bigl[\,|G_0(p)-G_1(p)|>35\,\bigr],
\end{equation}
and we keep $0.005\le\phi\le0.30$ on Bridge V2 and $0.004\le\phi\le0.5$ on the other datasets.
On Bridge V2 we further drop tasks that fold, wipe, open or close something, use drawers, put an object into a container, or only pick an object up, and keep at most $40$ trajectories per task.
On DROID we start from the $56{,}479$ successful episodes that have a language annotation and camera calibration, drop instructions with several movers, several steps, or articulated objects, keep at most $3$ episodes per (lab, instruction), and require the grasp to be held for at least $8$ steps.
On RT-1 we skip pick-only tasks; on Language-Table the task text is not used.

\noindent\textbf{Image geometry.}
Frames are processed at $640\times480$ ($912\times512$ for DROID).
DROID provides calibrated per-episode intrinsics; for the other datasets we assume $f_x=f_y=610$ and $(c_x,c_y)=(320,240)$.

\subsection{Measuring the move}
\label{app:real_measure}

\noindent\textbf{Segmentation and depth.}
SAM~3 segments both frames with text prompts: the robot arm and gripper, a fixed list of common object nouns (28 kitchen nouns, or 46 household nouns for DROID), generic fallbacks (``small object'', ``toy food'', ``kitchen item'', ``handheld object''), and the nouns of the task instruction.
We keep masks scoring at least $0.35$ ($0.20$ for task nouns) with an area between $300$~px and $5\%$ of the frame.
Depth Anything~3 predicts depth for $I_0$ and $I_1$ in a single call, so both frames share one scale.

\noindent\textbf{3D positions.}
For an object with mask $M$, let $(u,v)$ be the median pixel of $M$ and $z$ the median depth over $M$.
Its 3D position is
\begin{equation}
X
=
\frac{z}{g}\,K^{-1}\,[\,u,\;v,\;1\,]^{\top},
\end{equation}
where $K$ is the intrinsic matrix and $g$ the per-dataset scale factor (\cref{app:real_scale}).

\noindent\textbf{Matching objects across frames.}
A frame-0 mask and a frame-1 mask are match candidates if they share a class label, their HSV colour histograms correlate at least $0.5$, and their areas differ by less than a factor of $4$.
Pairs that share only a generic label must reach a correlation of $0.75$ and lie close to the predicted position.
Candidates implying more than $30$~cm of travel ($90$~cm for DROID) are refused.
Point tracks provide the main evidence: if the tracked endpoint of an object lies within $60$~px of a frame-1 mask, that mask is its match.
A class with exactly one instance in each frame is matched directly; otherwise, dense optical flow only breaks ties.

\noindent\textbf{Mover and landmarks.}
Let $\Delta=X^{(1)}-X^{(0)}$ be an object's displacement.
The mover $A$ must satisfy $\|\Delta\|\ge3$~cm and move at least $8$~px in the image; objects with $\|\Delta\|<2$~cm are candidate landmarks.
Pairs with more than two other moving objects are skipped, as are rotation tasks (flip, topple, turn upright).
Where the task text is used, the mover must be the object the task names.
A mover with more than $30\%$ of its after-frame mask inside the $25$-px-dilated arm mask is still held by the gripper and is rejected.

\noindent\textbf{Tracking certification.}
AllTracker tracks every pixel of $I_0$ through up to $48$ frames of the episode.
Let $\bar v$ be the median track displacement over the mover's before-frame mask and $\delta$ the mover's measured image displacement.
The pair is certified if
\begin{equation}
\cos(\bar v,\delta)>0.7
\quad\text{and}\quad
0.7\le\frac{\|\bar v\|}{\|\delta\|}<2 .
\end{equation}
The lower bound is $0.5$ in the certification step and is raised to $0.7$ by a later filter, which caught a wrong object being marked as the mover.

\subsection{Metric scale}
\label{app:real_scale}
For Language-Table, a homography fitted from recorded end-effector positions recovers table-plane geometry.
Other sources use dataset-level depth scaling checked against objects of known size; their absolute metric distances remain approximate.

\noindent\textbf{Per-dataset factors.}
Depth Anything~3's metric output is divided by one factor per dataset: $g=1.95$ (Bridge V2), $1.15$ (DROID), $0.61$ (RT-1), and $0.84$ (Berkeley UR5).
Factors are checked against objects of standard size $s_o$ (a can $12$~cm, a cup $9$~cm, an apple $8$~cm, \ldots): with the implied extent
\begin{equation}
e_g(o)
=
\frac{\max(w_o,h_o)\,z_o}{f\,g},
\end{equation}
where $w_o,h_o$ are the box width and height in pixels, a factor is scored by $\operatorname{median}_o\bigl|\log\bigl(e_g(o)/s_o\bigr)\bigr|$.
On DROID, a per-pair factor derived from the gripper's known depth was no better than a single constant.
On Bridge V2, the size check is confounded by toy food, so absolute centimetres there remain approximate.

\noindent\textbf{Language-Table table plane.}
For Language-Table we recover metric geometry from the robot's recorded end-effector positions.
A homography $H$ between the image and the table plane is fitted with RANSAC (inlier threshold $8$~px) to first-frame correspondences between the bottom-centre pixel of the arm mask and the recorded end-effector position, pooling at least $30$ correspondences from pairs taken with the same camera setup (identified by the position of the board edges in the image).
The median residual is $12.4$~px ($90$th percentile $14.7$~px), roughly $1$~cm.
Each object is placed on the plane at the bottom centre of its box raised by a quarter of the box height, and its motion perpendicular to the board is set to zero.

\noindent\textbf{Scale consistency.}
Every question states the size $c_{\mathrm{ref}}$ (in cm) of a reference object whose box spans $\pi_{\mathrm{ref}}$ pixels at its widest, and a move of $c_{\mathrm{mv}}$~cm whose image displacement is $\pi_{\mathrm{mv}}$ pixels.
When $c_{\mathrm{mv}}\ge5$~cm and $\pi_{\mathrm{mv}}\ge5$~px, we require the two implied pixels-per-centimetre to agree:
\begin{equation}
0.6
\le
\frac{\pi_{\mathrm{mv}}/c_{\mathrm{mv}}}{\pi_{\mathrm{ref}}/c_{\mathrm{ref}}}
\le
1.67 .
\end{equation}

\subsection{Question generation}
\label{app:real_questions}

\noindent\textbf{Landmarks.}
Landmarks are matched non-moving objects, excluding surfaces and fixtures (tables, counters, walls, knobs, faucets, \ldots), objects under $1000$~px ($500$~px for Language-Table, RT-1, and Berkeley UR5), objects that lose more than $60\%$ of their visible area in $I_1$ or are more than $25\%$ covered by the arm there, objects whose centre lies inside the box of a nearer marked object, and duplicate masks of one object.
Marks closer than $30$~px to an existing mark are dropped (larger objects first), leaving at most six landmarks.
Mark numbers are shuffled per pair, so the mover is not always mark~1.

\noindent\textbf{Distances.}
Each object gets an effective radius from its box,
\begin{equation}
r_o
=
\frac{w_o+h_o}{4}\cdot\frac{z_o}{f\,g},
\end{equation}
($w_o/2$ in table-plane units for Language-Table), and distances are measured between surfaces:
\begin{equation}
d_t(L)
=
\max\Bigl(0,\;\bigl\|X_L-X_A^{(t)}\bigr\|-r_L-r_A^{(t)}\Bigr),
\qquad t\in\{0,1\},
\end{equation}
where $X_L$ is the landmark's before-frame position and $X_A^{(t)}$ the mover's position in frame $t$.
Landmarks separated from the mover mainly in depth, $|z_L-z_A^{(1)}|>25$~cm and $|z_L-z_A^{(1)}|>0.7\,d_1(L)$, are dropped, because monocular depth is least reliable along that axis.
Unless stated otherwise, every comparison must clear a margin $m=3$~cm ($6$~cm for DROID, whose scenes are room-scale).

\noindent\textbf{Move text and scale reference.}
The move states the mover's displacement $\Delta$ along $x$ (``to the left/right'') and $y$ (``up/down''), rounded to whole centimetres, omitting components below $2$~cm; the depth component is not stated.
A second mover, if any, is described the same way.
The scale reference is the largest landmark whose extent $\max(w\,z/f_x,\,h\,z/f_y)/g$ lies between $3$ and $60$~cm (footprint width between $2$ and $60$~cm for Language-Table).

\noindent\textbf{Marks.}
The numbered version draws a red numbered badge on every marked object.
The colour-dot version draws only the objects a question mentions, with colours drawn at random per question; a dot never takes the palette colour nearest or second nearest to the object beneath it, nor any colour within $40$ CIELAB units of it.

\noindent\textbf{Flips.}
For each pair we first look for questions whose answer changes because of the move (below).
Only if none passes do we emit a no-change question (\cref{app:real_nochange}).

\subsubsection{Threshold}
Is the mover within $T$ of landmark $L$ after the move?
The answer is yes iff $d_1(L)<T$, with $T\in\{10,15,20,30\}$~cm.
We require a flip, $\mathbf{1}[d_0(L)<T]\ne\mathbf{1}[d_1(L)<T]$, and margins
\begin{equation}
|d_1(L)-T|\ge5~\mathrm{cm},
\qquad
|d_0(L)-T|\ge3~\mathrm{cm}.
\end{equation}
For DROID, $T\in\{15,20,30,50,75,100\}$~cm, $|d_1(L)-T|\ge\max(8~\mathrm{cm},0.15T)$, and $|d_0(L)-T|\ge5$~cm.

\subsubsection{Left Right}
Will the mover end up to the left of $L$, from the camera's viewpoint?
The answer is yes iff $x_A^{(1)}<x_L$.
We require the side to flip between the two frames and $|x_A^{(1)}-x_L|\ge m$ ($\ge\max(m,\,w_L/2)$ for DROID, with $w_L$ the landmark's metric width).

\subsubsection{Closer To}
Which of the listed landmarks will the mover end up closest to?
With at least three landmarks, let $L^\star=\arg\min_L d_1(L)$ and $L^{(2)}$ the second nearest after the move.
We require $\arg\min_L d_0(L)\neq L^\star$ and $d_1(L^{(2)})-d_1(L^\star)>2m$.
The options are $L^\star$ and up to three other landmarks, ordered by mark number.

\subsubsection{Closest Three}
Which three listed landmarks will the mover end up closest to?
With at least five landmarks, let $\mathcal{T}_t$ be the three nearest in frame $t$ and $L_{(k)}$ the $k$-th nearest after the move.
We require $\mathcal{T}_0\ne\mathcal{T}_1$ and $d_1(L_{(4)})-d_1(L_{(3)})>1.5m$.
The options are the first six landmarks by mark number; the answer is $\mathcal{T}_1$ (three letters).

\subsubsection{Count Near}
How many of the listed landmarks will end up within $T$ of the mover?
Let $\mathcal{P}$ be the first six landmarks by mark number (at least four) and
\begin{equation}
n_t(T)
=
\#\bigl\{L\in\mathcal{P}:\,d_t(L)<T\bigr\}.
\end{equation}
For $T\in\{15,20,25,30,40\}$~cm ($\{30,50,75,100\}$~cm for DROID), we require $n_0(T)\ne n_1(T)$, $0<n_1(T)<|\mathcal{P}|$, and $|d_1(L)-T|\ge3$~cm for every $L\in\mathcal{P}$ ($\ge\max(8~\mathrm{cm},0.15T)$ for DROID), and pick the $T$ whose $n_1$ is closest to $|\mathcal{P}|/2$.
The options are $n_1$ and up to three values from $\{n_1\pm1,\,n_1\pm2,\,n_0\}$ within $[0,|\mathcal{P}|]$.

\subsubsection{Depth Order}
Will the mover end up closer to the camera than $L$?
The answer is yes iff $z_A^{(1)}<z_L$.
Only compact landmarks qualify (box aspect ratio at most $3$, area at most $2\%$ of the frame, at least $20$~px from the border), and we require
\begin{equation}
|z_A^{(1)}-z_L|\ge15~\mathrm{cm},
\qquad
|z_A^{(0)}-z_L|\ge8~\mathrm{cm}.
\end{equation}
At most one no-change Depth Order question is kept per pair.
Depth Order questions come from Bridge V2 and DROID only, since depth differences on the top-down Language-Table board and on RT-1 are not reliable.

\subsubsection{Occlusion}
After the move, will $L$ be fully visible, or partly blocked?
Let $a_t(L)$ be the visible mask area of $L$ in frame $t$, and only consider landmarks at most $25\%$ covered by the arm in $I_1$.
\emph{Cover} (answer: partly blocked) requires the mover to end up in front of $L$ and to hide part of it:
\begin{equation}
\begin{gathered}
a_1(L)\le0.85\,a_0(L),
\qquad
z_A^{(1)}<z_L-1~\mathrm{cm},\\
\frac{\bigl|M_A^{(1)}\cap M_L^{(0)}\bigr|}{\bigl|M_L^{(0)}\bigr|}\ge0.25 .
\end{gathered}
\end{equation}
\emph{Reveal} (answer: fully visible) requires $L$ to be uncovered when the mover leaves: $a_1(L)\ge1.35\,a_0(L)$, with the boxes of $L$ and the mover overlapping by at least $10\%$ of $L$'s box before the move and at most $5\%$ after.
At most one Occlusion question is generated per pair.

\subsubsection{No-change questions}
\label{app:real_nochange}
For pairs without any flipped question, we ask a Threshold or Left Right question whose answer is the same before and after the move.
For Threshold, $T$ is drawn from $\{10,15,20,30,50\}$~cm before looking at the geometry, and both $d_0(L)$ and $d_1(L)$ must lie on the same side of $T$ by at least $m$; for Left Right, $|x_A^{(t)}-x_L|>5$~cm on the same side in both frames.
Half of these questions state the false claim (``more than $T$ away'', or the wrong side), so that the answer is no.

\subsubsection{Named-object questions}
The 103 named-object questions come from Bridge V2 and DROID and use the same geometry with object names instead of marks.
Qwen3.5-9B names each object from an image crop; names with generic heads (``object'', ``item'', ``container'', \ldots), robot parts, and cables are rejected.
Qwen3.5-9B then counts the visible objects that fit each name, and a name that fits no object, or fits two or more objects without a distinguishing qualifier, is rejected, as is the same object appearing under two names.
Threshold margins are $|d_1(L)-T|\ge4$~cm; Closest Three requires a $3$~cm gap; and Occlusion counts a cover from $10\%$ overlap.
Every direction carries ``(directions as seen from the camera)''.

\subsection{Filtering}
\label{app:real_filters}
Most filters below were added after a human review round found a failure of that kind; each was then applied to the whole pool.
An item is removed if any of the following holds:
\begin{itemize}
\item the mover is elongated (box aspect ratio above $3.5$), covers less than $1000$~px, shrinks below $30\%$ of its visible area after the move, or has a depth change above $60$~cm;
\item the mover ends in the gripper, its end point is not a visible object, or the object never moved (its appearance at the old location still correlates above $0.6$ with the before frame);
\item a mark falls on the robot or a person (checked on crops with Moondream~3 and on the full image with Qwen3.5-9B), or a mark moves with the arm;
\item two marks share one object: one mask is nested in another (containment above $0.7$ at an area ratio of at least $0.33$), covers at least $40\%$ of it, or the two are touching parts of one fixture;
\item a stated direction contradicts the tracked motion, or the stated magnitude is far below or above the tracked one;
\item the stated move is below $5$~cm in total, or fails the scale-consistency check (\cref{app:real_scale});
\item a 3D distance is much smaller than its image-plane counterpart ($d<0.6\,d_{\mathrm{plane}}$ for $d_{\mathrm{plane}}>8$~cm), which indicates depth that shrank a distance;
\item a ranking margin is thin (Closest Three: $d_1(L_{(4)})<1.15\,d_1(L_{(3)})$; Closer To: gap below $5$~cm), or a counted landmark lies within $25\%$ of $T$ from the threshold;
\item a threshold exceeds $85\%$ of the scene's extent;
\item a container was knocked over during the move (tall in $I_0$, lying in $I_1$);
\item for named objects, a name check above fails.
\end{itemize}

\subsection{Splits and balance}
\label{app:real_split}
Test and training share no image pair, and the test set has one question per image pair.
Test items must not be borderline: we require $|d-T|/T\ge0.25$ and a margin of at least $5$~cm.
Selection targets an equal number of questions per category, and within each category caps the share of no-change questions and of any single answer at $55\%$ and the share of Left Right questions answered correctly by the shortcut ``the object ends up on the side it moved toward'' at $65\%$; yes/no answers are then trimmed to at most $55{:}45$.
In the final test set, the shortcut answers $105$ of $193$ Left Right questions correctly.
A text-only model, Qwen3.5-9B without the image (temperature $0$, thinking off), answers every item in the original and in the reversed option order; items it answers correctly in both orders (for questions with more than two options, in the original order) are kept but selected only after all other items of their category.

\subsection{Training split}
\label{app:real_train}
The image pairs not used for testing yield a training split of $8{,}575$ questions on $5{,}367$ pairs, built with the same rules.
Its categories are imbalanced, because flipped Occlusion, Depth Order, Closest Three, and Count Near questions are rare in the mined pairs: Threshold $6{,}831$, Depth Order $669$ (almost all no-change), Left Right $448$, Count Near $317$, Closer To $209$, Closest Three $81$, and Occlusion $20$.
An extended version of $9{,}748$ questions adds questions generated with halved margins on the same training pairs, keeping only categories whose answers do not depend on absolute scale (mostly Closer To and Closest Three); all nine of the thinnest-margin Closer To and Closest Three questions we reviewed were correct.

\section{Implementation Details}
\subsection{Experimental Setups}
\label{supp:exp_setup}
\noindent\textbf{Proprietary-model inference.}
We access GPT-6 Astra and GPT-5.6 Luna through their APIs, and Claude-Fable-5.1 through Claude Code in headless mode, with one fresh process per question.
Each question is submitted as a fresh single-turn request containing only the image and prompt, without a system prompt, conversation history, or tools; for Claude-Fable-5.1, about 800 tokens of fixed Claude Code context remain in each request.
We score the first response, retry requests only on errors, and leave temperature unspecified.
To limit API cost, GPT-6 Astra is evaluated on 50\% of each benchmark, while Claude-Fable-5.1 and GPT-5.6 Luna are each evaluated on 20\%, with questions sampled by category.

\noindent\textbf{Reasoning settings.}
Because reasoning cannot be fully disabled for the proprietary models, modes 1A and 2A use the lowest supported reasoning effort, while 1B uses the highest (\emph{low} and \emph{max} for Claude).
The prompts are identical in 1A and 1B; only the reasoning-effort setting changes.

\noindent\textbf{Open-source models and inference.}
We evaluate nine open-source VLMs from four families:
Qwen3-VL-8B and -32B~\cite{bai2025qwen3},
Qwen3.5-9B~\cite{qwen35},
InternVL3.5-8B and -30B-A3B~\cite{wang2025internvl3},
LLaVA-OneVision-1.5-8B~\cite{an2025llavaov} and -2-8B~\cite{an2026llavaov},
and Gemma-4-12B and -31B~\cite{team2026gemma}.
Each question is evaluated under three answer-order shuffles, and we report mean accuracy.
In mode 1B, models are prompted to reason step by step before answering, with a budget of 4{,}096 tokens.

\noindent\textbf{Human-evaluation protocol.}
The twelve participants include three paper authors.
The 240 FlowBench-Real-Test questions cover all seven categories and are presented on pages of 35 questions.
Participants receive the same image, motion instruction, and question as models in mode 1A.
Their responses yield 403 answers, with most questions answered by two participants.
For questions with two responses, agreement ranges from 65--82\% across categories; when participants agree, their answers are correct 73--100\% of the time.
Participants additionally provide 120 answers on FlowBench-Sim-Test.

\subsection{Fine-tune Settings}
\label{supp:fine-tune}
We fine-tune three open-source VLMs, Qwen3-VL-8B-Instruct, Qwen3.5-9B and LLaVA-OneVision-2-8B-Instruct, with an identical recipe on FlowBench-Sim-Train in setting 1A (before image with colour Set-of-Marks and the action as input, answer letter as target; 317{,}019 training examples). We use parameter-efficient fine-tuning with LoRA~\cite{hu2021lora} (rank $r{=}16$, $\alpha{=}32$, dropout $0.05$) on the query, key, value and output projections and the gate, up and down MLP projections of the language model; the vision encoder and the vision--language projector stay frozen. This adds 43.6M trainable parameters for Qwen3-VL-8B and LLaVA-OneVision-2-8B (0.5\% of all parameters) and 29.1M for Qwen3.5-9B (0.3\%), whose hybrid architecture has these attention projections only in its 8 full-attention layers (its Gated DeltaNet layers are not adapted). Qwen3.5-9B is trained and evaluated in its non-thinking (instruct) mode. The loss is the cross-entropy on the assistant answer tokens only; prompt and image tokens are masked. We train for 2 epochs (19{,}814 optimizer steps) with AdamW ($\beta_1{=}0.9$, $\beta_2{=}0.999$, no weight decay), a peak learning rate of $10^{-4}$ with linear warm-up over the first 3\% of steps followed by linear decay, gradient clipping at 1.0, and an effective batch size of 32 (micro-batch 1 per GPU, 8 gradient-accumulation steps, 4 GPUs). Training uses bfloat16 mixed precision, gradient checkpointing and a maximum sequence length of 2{,}048 tokens, with images processed at each model's default resolution. We compute the validation loss on 5{,}000 held-out examples every quarter epoch and keep the checkpoint with the lowest loss, which was the final one for all three models. Each model is trained on 4 NVIDIA A40 (48\,GB) GPUs, taking 44.2\,h (Qwen3-VL-8B), 67.7\,h (Qwen3.5-9B) and 40.4\,h (LLaVA-OneVision-2-8B) of wall-clock time, i.e.\ about 177, 271 and 161 GPU-hours. At test time, each fine-tuned model receives the same prompt format it was trained on and is decoded greedily. All training uses Hugging Face Transformers and PEFT.

\subsection{Scoring}
We report accuracy; a question counts as correct only if the predicted letters exactly match the answer, so Closest Three requires all three letters. We read the answer from the end of each reply, and missing or unreadable replies count as wrong. Chance is the mean over questions of one over the number of options: 35.8\% on FlowBench-Sim and 37.6\% on FlowBench-Real.

\section{Extended Results and Analysis}
\label{app:extra_results}

\subsection{Main Results}
This section complements \cref{sec:exp_results} with the details behind \cref{tab:results_1a}.

\noindent\textbf{Scale and reasoning effort.}
Scale helps open-source models little: Qwen3-VL-32B gains only 1.4 points on FlowBench-Sim and 1.7 on FlowBench-Real over Qwen3-VL-8B, and InternVL3.5-30B-A3B is 1.5 points below InternVL3.5-8B on Real. Max reasoning effort raises the Real accuracy of GPT-6 Astra and Claude-Fable-5.1 to 89.9\% and 88.9\%. On FlowBench-Sim at low effort, both models stay near or below chance on Closest Three (23.3\% and 33.3\%) and Count Near (19.0\% and 30.8\%).

\noindent\textbf{Categories.}
Closest Three is the hardest category for open-source models, which score at most 17.0\% on Sim and 19.0\% on Real, against chance levels of 5.0\% and 7.0\%. On Sim Count Near, seven of the nine open-source models score below chance, and on the binary Real categories most stay near 50\%; four score exactly 50.0\% on Depth Order. Left Right is the easiest category, with up to 86.2\% on Sim and 64.1\% on Real for open-source models.

\noindent\textbf{Models versus humans.}
Pooled human accuracy is 77.2\% on FlowBench-Real (95\% Wilson interval 72.8--81.0\%) and 64.2\% on FlowBench-Sim (55.3--72.2\%). On Real, humans are most accurate on Closer To (87.8\%) and Closest Three (87.2\%) and least accurate on Count Near (63.5\%) and Depth Order (70.4\%), which require metric judgments from a single image. On Sim, humans lead every base model, including GPT-6 Astra at max effort (60.2\%), but with 120 answers this lead over the proprietary models lies within the interval. Humans and models answered different subsets of the questions, so these comparisons are indicative.

\noindent\textbf{Sim-to-real transfer by category.}
Fine-tuning on FlowBench-Sim-Train raises FlowBench-Real accuracy by 16.9 points for Qwen3-VL-8B, 19.9 for LLaVA-OV-2-8B, and 25.0 for Qwen3.5-9B. Transfer is strongest on Closest Three, which rises from at most 12.8\% to 58.0--62.0\%, and on Near Object (+27.8 to +31.1 points). In contrast, all three models lose accuracy on Real Count Near, and only Qwen3.5-9B improves on Occlusion. The fine-tuned models exceed both humans and frontier models on Sim, but still trail them on Real.

\subsection{Imagining versus Observing for Proprietary Models}
\label{app:gpt_modes}

\begin{table}[t]
\centering
\small
\setlength{\tabcolsep}{1.6pt}
\begin{tabular}{llcc|cc|cc}
\toprule
 & & & & \multicolumn{2}{c|}{FlowBench-Sim} & \multicolumn{2}{c}{FlowBench-Real} \\
Model & Set. & Img. & Eff. & Avg & Cat. & Avg & Cat. \\
\midrule
GPT-6 Astra & 1A & $I_0$ & low & 54.6 & 57.6 & 88.4 & 87.6 \\
 & 1B & $I_0$ & max & \textbf{60.2} & \textbf{64.4} & 89.9 & 89.2 \\
 & 2B & $I_1$ & max & 56.3 & 60.9 & \textbf{93.0} & \textbf{92.0} \\
\midrule
GPT-5.6 Luna & 1A & $I_0$ & low & 47.6 & 52.2 & 65.0 & 63.2 \\
 & 2A & $I_1$ & low & 42.7 & 47.7 & 60.4 & 60.9 \\
\midrule
\multicolumn{4}{l}{\textit{$\Delta$ Astra 2B$-$1B}} & \textcolor{gray}{$-$3.9} & \textcolor{gray}{$-$3.4} & +3.1 & +2.8 \\
\multicolumn{4}{l}{\textit{$\Delta$ Astra 1B$-$1A}} & +5.6 & +6.7 & +1.5 & +1.6 \\
\multicolumn{4}{l}{\textit{$\Delta$ Luna 2A$-$1A}} & \textcolor{gray}{$-$4.9} & \textcolor{gray}{$-$4.5} & \textcolor{gray}{$-$4.6} & \textcolor{gray}{$-$2.3} \\
\bottomrule
\end{tabular}
\caption{GPT models when imagining the outcome from the before image $I_0$ (1A, 1B) and when seeing the after image $I_1$ (2A, 2B). Eff.: reasoning effort; 1A and 2A use the lowest, 1B and 2B the highest; 2B is the after-image counterpart of 1B. GPT-6 Astra answers 50\% and GPT-5.6 Luna 20\% of the questions of each benchmark, the same samples as \cref{tab:results_1a}; Astra was not run in 2A and Luna not at max effort. Avg is per question and Cat.\ the mean of the category accuracies. Bold: best per column; drops in grey.}
\label{tab:gpt_modes}
\end{table}

\cref{tab:gpt_modes} compares GPT-6 Astra and GPT-5.6 Luna when they must imagine the outcome from the before image and when they see the after image, on the same question samples as \cref{tab:results_1a}. GPT-6 Astra's after-image control was run at max reasoning effort (2B), so we compare it with its max-effort before-image run (1B). GPT-5.6 Luna was run at low effort only, so its comparison is 2A versus 1A.

\noindent\textbf{FlowBench-Real.}
Seeing the outcome helps GPT-6 Astra: at max effort, its accuracy rises by 3.1 points (89.9\% to 93.0\%), with gains in every category except Depth Order (83.3\% to 81.9\%). Imagining the motion thus remains a measurable cost even for the strongest model with reasoning (\cref{sec:analysis_bottleneck}), while raising the effort in the before-image setting adds only 1.5 points (1A to 1B). GPT-5.6 Luna behaves differently: it is 4.6 points less accurate when shown the outcome (65.0\% to 60.4\%), mainly on Closer To (77.8\% to 53.3\%) and Near Object (69.7\% to 57.6\%), and it stays at 50.0\% on Depth Order in both settings. Its errors therefore already arise when reading spatial relations from a single image, so its imagination gap cannot be measured.

\noindent\textbf{FlowBench-Sim.}
Seeing the outcome lowers the accuracy of both models, by 3.9 points for GPT-6 Astra at max effort and by 4.9 points for GPT-5.6 Luna. For GPT-6 Astra, the largest drops are on Near Object (61.8\% to 50.9\%), Feasibility (65.5\% to 55.2\%), and Closer To (75.0\% to 68.2\%). This is consistent with \cref{sec:analysis_sim}: on Sim, the bottleneck is judging metric relations in the rendering itself rather than imagining the motion. Reasoning effort also helps more on Sim than on Real (+5.6 vs.\ +1.5 points from 1A to 1B). Claude-Fable-5.1 shows the same pattern on Sim, with 51.7\% in 2A against 56.6\% in 1A.

\noindent\textbf{Share of errors due to imagination.}
On FlowBench-Real, GPT-6 Astra at max effort and Claude-Fable-5.1 at low effort answer 93.0\% and 90.3\% of the after-image questions correctly. Imagining the motion therefore accounts for 31\% and 34\% of their before-image errors, respectively.

\noindent\textbf{Text reasoning per category.}
On FlowBench-Real, max reasoning effort brings Claude-Fable-5.1 to its 2A accuracy on Closest Three (96.3\%) and Count Near (82.1\%), which can be solved by enumerating objects and distances in text, but leaves it below 2A on Closer To ($-3.0$), Left Right ($-4.5$), and Depth Order ($-6.9$); on Depth Order, reasoning even lowers its accuracy from 82.8\% to 79.3\%. GPT-6 Astra does not show this split: at max effort it remains below its after-image accuracy in six of seven categories, from 1.5 points on Closest Three to 6.6 on Near Object, and exceeds it only on Depth Order (+1.4). The per-category effect of reasoning thus differs between models, and with 16 to 105 questions per category we do not draw conclusions from it.

\subsection{Why is FlowBench-Sim Hard for Frontier Models?}
\label{app:sim_hard}

This section details the analysis summarized in \cref{sec:analysis_sim}.

\noindent\textbf{Size of the gap.}
At low effort, GPT-6 Astra scores 54.6\% on FlowBench-Sim against 88.4\% on FlowBench-Real, and Claude-Fable-5.1 56.6\% against 85.3\%. The smaller GPT-5.6 Luna drops by 17.4 points (47.6\% vs.\ 65.0\%). Open-source models differ by at most 6.2 points between the two benchmarks, because they are close to chance on both (\cref{tab:results_1a}).

\noindent\textbf{The gap lies in metric distances.}
Across the six categories shared by the two benchmarks, the largest drops involve distances between objects. From Real to Sim, GPT-6 Astra drops by 72.1, 63.3, and 27.6 points on Closest Three, Count Near, and Near Object, and Claude-Fable-5.1 by 51.9, 44.2, and 32.4. Both models remain above 90\% on Left Right, which only compares positions in the image plane. Seeing the outcome does not help either: on Sim Count Near, Claude-Fable-5.1 answers only 11.5\% of the static 2A questions correctly, against 30.8\% in 1A, and all proprietary models are less accurate on Sim when shown the outcome (\cref{app:gpt_modes}). Estimating distances in a single rendering is thus itself a bottleneck, before any imagination is needed.

\noindent\textbf{Two candidate explanations.}
First, \emph{scale}: FlowBench-Sim uses room-scale scenes, camera-axis motions of up to about a meter, and distance thresholds of 0.35--1.25\,m, with few cues of absolute size. FlowBench-Real uses motions of a few to a few tens of centimeters in camera-relative image directions and states a visible object's width as a scale reference (\cref{sec:real}). Second, \emph{possible exposure}: FlowBench-Real is built from public robot datasets~\citep{walke2023bridgedata,khazatsky2024droid,lynch2023interactive,brohan2023rt1} that may appear in the training data of frontier models, possibly including the later frames we use as outcomes, whereas FlowBench-Sim motions are synthesized by our planner and cannot have been seen.

\noindent\textbf{Humans also find Sim harder.}
Humans score 13.0 points lower on Sim than on Real (64.2\% vs.\ 77.2\%, \cref{tab:results_1a}). Their largest drops are on the same distance-based categories as those of the frontier models: Closest Three falls from 87.2\% to 25.0\% and Near Object from 81.0\% to 46.7\%, whereas Left Right stays at 86.7\% on both. Because our participants cannot have seen the data before, their drop cannot come from exposure: judging metric distances between small, distant objects in room-scale renderings without a size reference is hard for humans too, which supports the scale explanation. Exposure may still contribute to the larger drop of the frontier models (17--34 points, against 13 for humans). The per-category human samples on Sim are small (11--17 answers), so these category-level comparisons are indicative only.

\noindent\textbf{Testing the explanations.}
Both explanations can be tested directly: exposure by evaluating on newly recorded videos or on episodes released after the models' training cutoff, and scale by adding a reference object's size to the Sim prompts.

\subsection{Do Models Answer as if Nothing Moved?}
\label{app:change}

\begin{table}[t]
\centering
\small
\setlength{\tabcolsep}{2.2pt}
\begin{tabular}{llc|cccc}
\toprule
\multicolumn{3}{l|}{\textbf{FlowBench-Real}} & Chg. & Unchg. & Gap & Gap$_\text{mix}$ \\
\midrule
GPT-6 Astra & 1A & low & 87.3 & 91.2 & +3.8 & +8.0 \\
 & 1B & max & 88.9 & 92.5 & +3.7 & +6.3 \\
 & 2B & max & 92.9 & 93.2 & +0.3 & +0.1 \\
GPT-5.6 Luna & 1A & low & 67.3 & 59.0 & $-$8.3 & $-$5.3 \\
 & 2A & low & 62.2 & 55.7 & $-$6.4 & $-$6.2 \\
\midrule[\heavyrulewidth]
\multicolumn{3}{l|}{\textbf{FlowBench-Sim}} & Chg. & Unchg. & Gap & Stale (chance) \\
\midrule
GPT-6 Astra & 1A & low & 56.0 & 45.9 & $-$10.1 & 29.7 (27.5) \\
 & 1B & max & 61.3 & 52.7 & $-$8.6 & 24.1 (26.7) \\
 & 2B & max & 55.1 & 60.8 & +5.7 & 47.5 (26.8) \\
GPT-5.6 Luna & 1A & low & 43.3 & 57.7 & +14.4 & 44.4 (27.8) \\
 & 2A & low & 39.2 & 53.8 & +14.7 & 51.3 (28.2) \\
\bottomrule
\end{tabular}
\caption{Accuracy (\%) on questions whose answer the motion changes (Chg.) or does not change (Unchg.). Gap: Unchg.$-$Chg., so a positive gap means lower accuracy when the scene changes. Gap$_\text{mix}$ pools only the categories that contain both kinds of question (Depth Order, Left Right, and Near Object on Real; on Sim, every category does). Stale: share of wrong answers on changed multi-choice questions that equal the before-motion answer, with its chance level. Same question samples as \cref{tab:gpt_modes}; Feasibility is excluded because it has no before state.}
\label{tab:change}
\end{table}

Humans are less accurate when the motion changes the answer and tend to give the answer of the unchanged scene (\cref{sec:analysis_human}). \cref{tab:change} tests whether GPT models show the same tendency, splitting each benchmark into questions whose answer changes after the motion and questions whose answer does not. On FlowBench-Real, unchanged questions occur only in Depth Order, Left Right, and Near Object, so we compare the two groups within these categories (Gap$_\text{mix}$).

\noindent\textbf{FlowBench-Real.}
When it must imagine the outcome, GPT-6 Astra is 8.0 points less accurate on changed than on unchanged questions at low effort and 6.3 points at max effort. The gap comes mostly from Near Object (79.3\% vs.\ 97.9\%) and Depth Order (65.0\% vs.\ 80.8\%). It disappears when the model sees the outcome (+0.1 in 2B), so the deficit is specific to imagining the motion: like humans, though to a lesser degree, GPT-6 Astra partly answers as if nothing moved. GPT-5.6 Luna shows the opposite pattern and is more accurate on changed questions, in line with \cref{app:gpt_modes}, where its errors already arise when reading the scene. Since FlowBench-Real stores only after-motion distances for multi-choice categories, the stale rate cannot be computed there.

\noindent\textbf{FlowBench-Sim.}
The picture is mixed. GPT-6 Astra is more accurate on changed questions when imagining ($-10.1$ and $-8.6$), and its wrong answers equal the before-motion answer at about the chance rate. One explanation is a prior that answers change, since three quarters of the Sim questions with a before state are changed by design (\cref{app:planning_algorithms}). GPT-5.6 Luna is less accurate on changed questions (+14.4), and its wrong answers equal the before-motion answer well above chance (44.4\% vs.\ 27.8\%). GPT-6 Astra's high stale rate in 2B (47.5\%) rests on 61 wrong answers and is hard to interpret.

\noindent\textbf{Motion length.}
On FlowBench-Real, motions longer than 15\,cm are harder in 1A, by 4.4 points for GPT-6 Astra and 8.9 points for GPT-5.6 Luna. The difference shrinks with max reasoning effort for GPT-6 Astra ($-1.6$) and disappears when GPT-5.6 Luna sees the outcome (+0.4). GPT-6 Astra remains 3.0 points less accurate on long motions even in 2B, so long motions also come with scenes that are harder to read.

\noindent\textbf{Caveats.}
Before-motion answers are recovered from the simulator metadata for 74 of 80 changed Closer To, 66 of 79 Closest Three, and 61 of 84 Count Near questions on Sim; the remaining questions are excluded from the stale rate. GPT-5.6 Luna answers only 20\% of the questions (123 on Sim, 217 on Real), so its estimates are coarse.

\subsection{World-Model-Assisted Answering}
\label{app:wm}

\begin{table}[t]
\centering
\small
\setlength{\tabcolsep}{4pt}
\begin{tabular}{lr|cc|c}
\toprule
Category & $n$ & Direct & Imagine & $\Delta$ \\
\midrule
Closer To & 94 & 38.3 & 27.7 & \textcolor{gray}{$-$10.6} \\
Closest Three & 112 & 12.5 & 8.9 & \textcolor{gray}{$-$3.6} \\
Contact Wall & 19 & 63.2 & 63.2 & 0.0 \\
Count Near & 110 & 50.0 & 31.8 & \textcolor{gray}{$-$18.2} \\
Feasibility & 106 & 76.4 & 76.4 & 0.0 \\
Left Right & 116 & 90.5 & 90.5 & 0.0 \\
Near Object & 117 & 61.5 & 60.7 & \textcolor{gray}{$-$0.9} \\
Occlusion & 40 & 57.5 & 55.0 & \textcolor{gray}{$-$2.5} \\
\midrule
Avg & 714 & 55.7 & 50.7 & \textcolor{gray}{$-$5.0} \\
Cat-avg & & 56.2 & 51.8 & \textcolor{gray}{$-$4.5} \\
\bottomrule
\end{tabular}
\caption{PSI-0.6 on FlowBench-Sim when answering directly from $(I_0,t_m,t_q)$ and when it first imagines the flow and the outcome frame (imagine, then answer), with the same checkpoint. Accuracy in \%; drops in grey.}
\label{tab:wm}
\end{table}

This section details the study summarized in \cref{sec:analysis_wm}.

\noindent\textbf{Inference.}
In imagine-then-answer mode (\cref{fig:wm_vlm}), PSI-0.6 first predicts the optical flow $\hat{F}_{01}$ of the described motion from $(I_0,t_m)$, then generates the outcome frame $\hat{I}_1$ from the flow, and finally answers $t_q$ from the imagined outcome. During decoding, it generates the patches with the highest motion probability first and strengthens the motion description by combining text-conditioned and image-only logits with weights $+2$ and $-1$. Each answering mode uses its own training recipe: the PSI-0.6 rows of \cref{tab:results_1a} report the model trained for direct answering, whereas \cref{tab:wm} compares the two modes of the model trained for imagine-then-answer.

\noindent\textbf{Per-category results.}
With the same checkpoint, imagining before answering lowers accuracy from 55.7\% to 50.7\% (\cref{tab:wm}). The drop concentrates on Count Near ($-18.2$) and Closer To ($-10.6$), while Contact Wall, Feasibility, and Left Right are unchanged and Near Object barely moves ($-0.9$).

\noindent\textbf{Diagnosis.}
The pattern points to the quality of the imagined intermediates rather than to visual imagination itself. The categories that degrade most require every marked object to keep its identity and precise position in $\hat{I}_1$: Count Near compares the mover with six listed objects, and Closer To with up to four landmarks. Coarse relations such as Left Right survive imperfect renderings. Generated frames often violate this requirement: in \cref{fig:wm_vlm}, the dot marking the garbage can is missing from the predicted frame, so the question can no longer be grounded in it. Since seeing the true outcome helps strong models on FlowBench-Real (\cref{sec:analysis_bottleneck}), world models that preserve the details a question depends on should make visual imagination useful for CSI.

\subsection{Human Performance Analysis}
\label{app:human}

This section details the analysis summarized in \cref{sec:analysis_human}. Humans and models answered different subsets of questions, so the comparisons are indicative.

\noindent\textbf{Relations versus metric thresholds.}
Humans answer relational questions, which compare objects with one another (Closer To, Closest Three, Left Right), correctly 87.2\% of the time, but threshold questions that require a metric distance (Count Near, Near Object) only 71.3\% of the time. Frontier models drop less, from 94.2\% to 85.6\% for GPT-6 Astra and from 89.5\% to 80.6\% for Claude-Fable-5.1 at low effort, so their advantage over humans is largest on metric judgments (14.3 and 9.3 points, against 7.0 and 2.3 on relations). Two participants also reported that judging centimeter thresholds from a single image felt unnatural.

\noindent\textbf{Humans answer as if nothing moved.}
Humans are correct on 84\% of the questions whose answer the motion does not change, but on only 68.5\% of those whose answer it changes. On Depth Order questions whose answer changes, they score 54\%, close to chance, and every error is the answer that the before image would give. Long motions (over 15\,cm) are also harder than short ones (72\% vs.\ 82\%). Human errors thus come mainly from failing to update the scene, \ie from imagination itself. GPT-6 Astra shows a weaker form of this tendency, which disappears when it sees the outcome (\cref{app:change}).

\noindent\textbf{Human errors are not label errors.}
We manually re-checked every question that both participants answered incorrectly and found the label correct in all cases. On 91\% of the questions with two answers, at least one participant was correct.

\noindent\textbf{Implications.}
On FlowBench-Real, humans and frontier models have different strengths: the models are better at the metric distance judgments that humans find hard, while human errors come mostly from not updating the scene. The models reach this without explicit 3D inputs or geometric modules, which complements approaches that add such inductive biases~\cite{hong20233d,chen2024spatialvlm,cheng2024spatialrgpt}; our evaluation cannot tell whether scale, data, or other factors are responsible.

\section{Discussion on Limitations}
\label{app:limitations}

\noindent\textbf{Perception and imagination.}
1A accuracy combines reading the scene with imagining the motion. The 2A control separates the two only where the static task is easy: on FlowBench-Sim, static metric questions are themselves hard, so 1A there also measures metric perception. We therefore report 2A alongside 1A and draw conclusions about imagination from their gap, mainly on FlowBench-Real.

\noindent\textbf{Measured geometry.}
FlowBench-Real answers rely on estimated depth, masks, and tracks; margins guard against measurement noise, but absolute distances remain approximate.

\noindent\textbf{Evaluation coverage.}
For cost reasons, proprietary models answer 20--50\% of the questions, on samples that differ between models, and the human baseline covers different subsets of 120 and 403 answers, with small per-category samples. Its centimeter-based instructions may also disadvantage humans, so human accuracy is a reference point rather than an upper bound.

\noindent\textbf{Possible exposure.}
FlowBench-Real is built from public robot datasets that frontier models may have seen during training.

\noindent\textbf{Scope.}
Motions are single rigid translations of one object in indoor and tabletop scenes, without rotations, deformations, or multi-step interactions. Our world-model study covers one model and one inference procedure, and does not yet test whether more faithful intermediates, such as the simulator's true flow and outcome frame, would recover accuracy.

\section{Visualizations}
\label{app:visualizations}

\begin{figure*}[p]
\centering
\includegraphics[width=\textwidth,height=\dimexpr\textheight-3\baselineskip\relax,keepaspectratio]{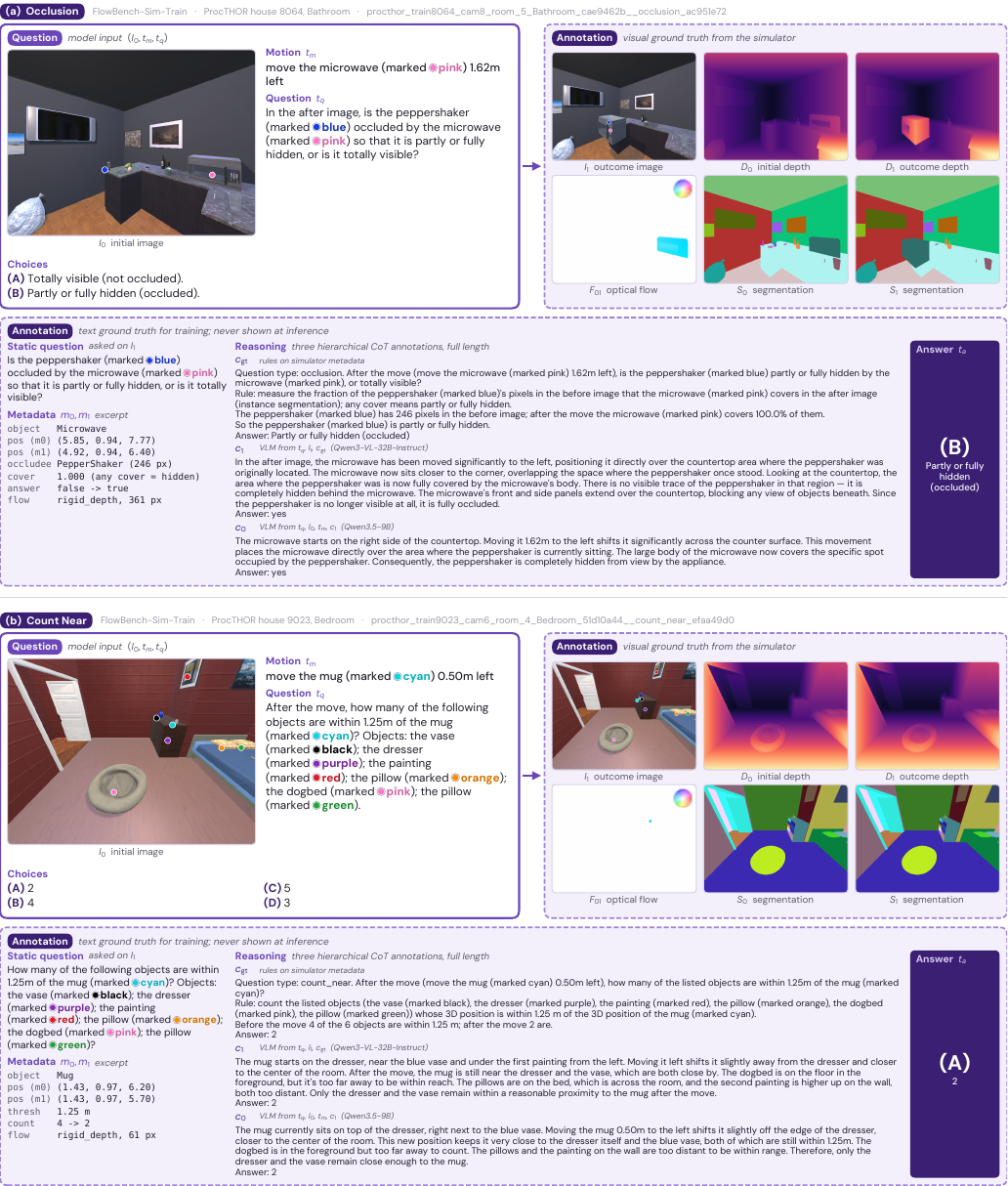}
\caption{Visualization of FlowBench-Sim-Train samples: (a) Occlusion and (b) Count Near.}
\label{fig:vis_sim_1}
\end{figure*}

\begin{figure*}[p]
\centering
\includegraphics[width=\textwidth,height=\dimexpr\textheight-3\baselineskip\relax,keepaspectratio]{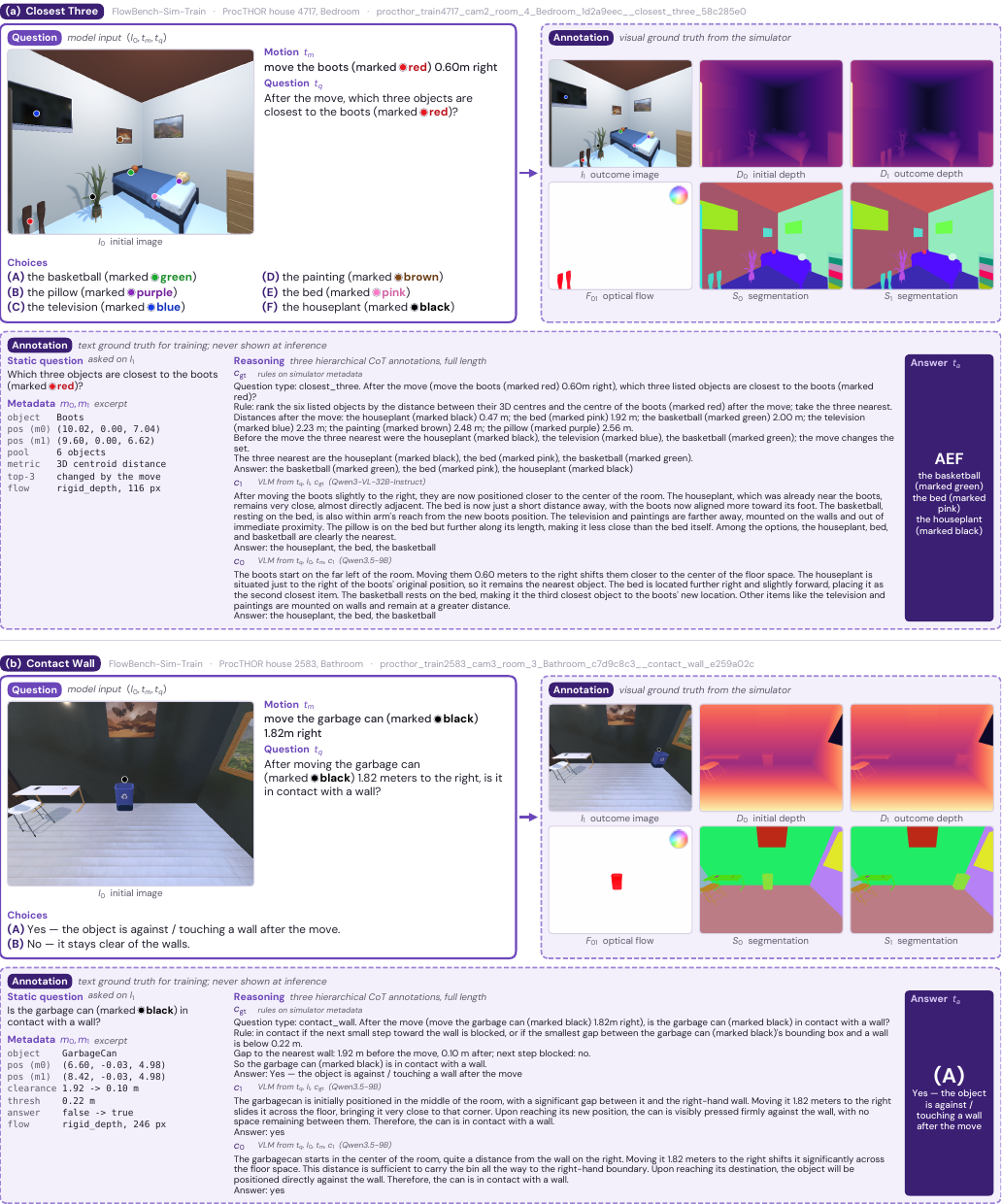}
\caption{Visualization of FlowBench-Sim-Train samples: (a) Closest Three and (b) Contact Wall.}
\label{fig:vis_sim_2}
\end{figure*}

\begin{figure*}[p]
\centering
\includegraphics[width=\textwidth,height=\dimexpr\textheight-3\baselineskip\relax,keepaspectratio]{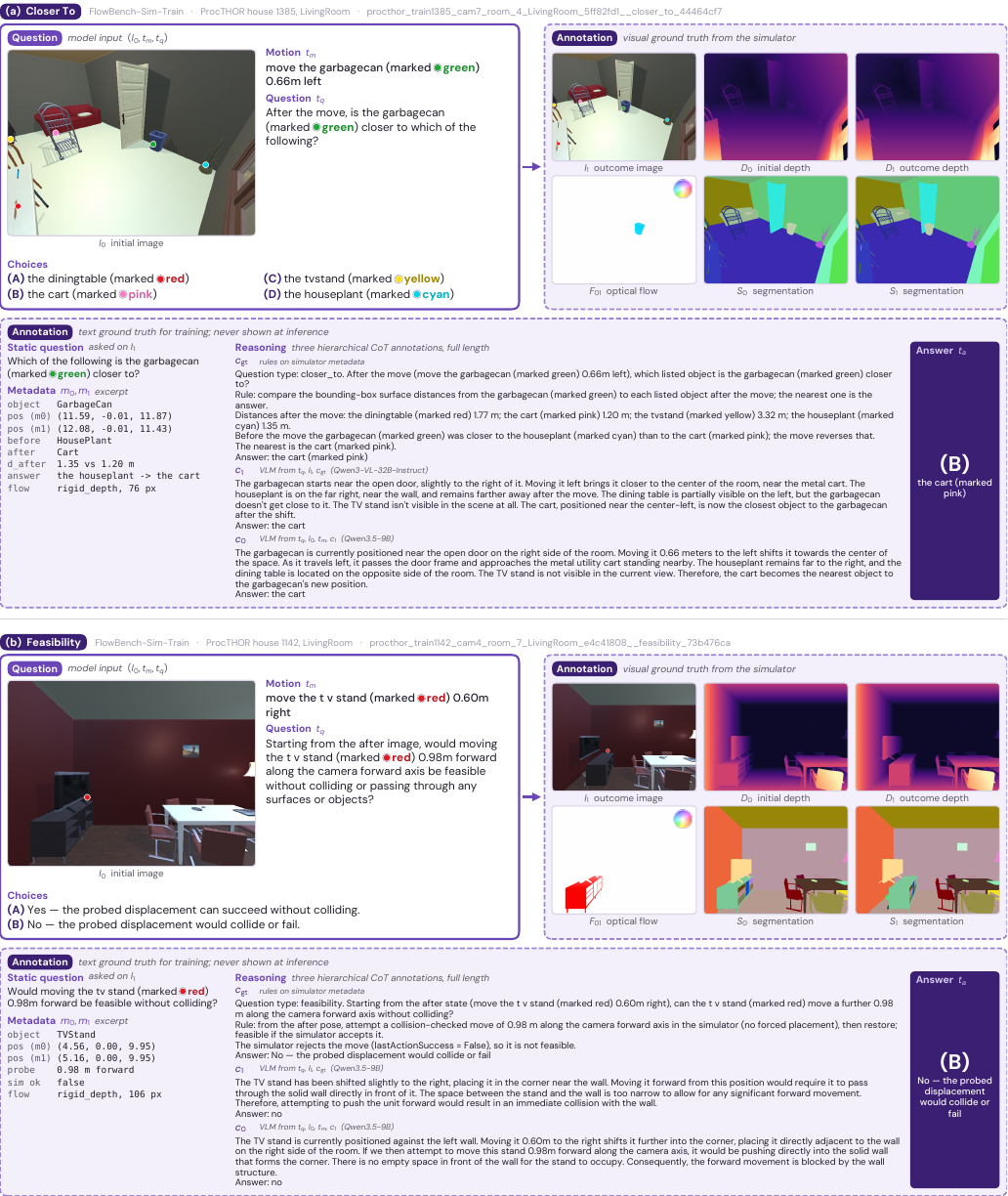}
\caption{Visualization of FlowBench-Sim-Train samples: (a) Closer To and (b) Feasibility.}
\label{fig:vis_sim_3}
\end{figure*}

\begin{figure*}[p]
\centering
\includegraphics[width=\textwidth,height=\dimexpr\textheight-3\baselineskip\relax,keepaspectratio]{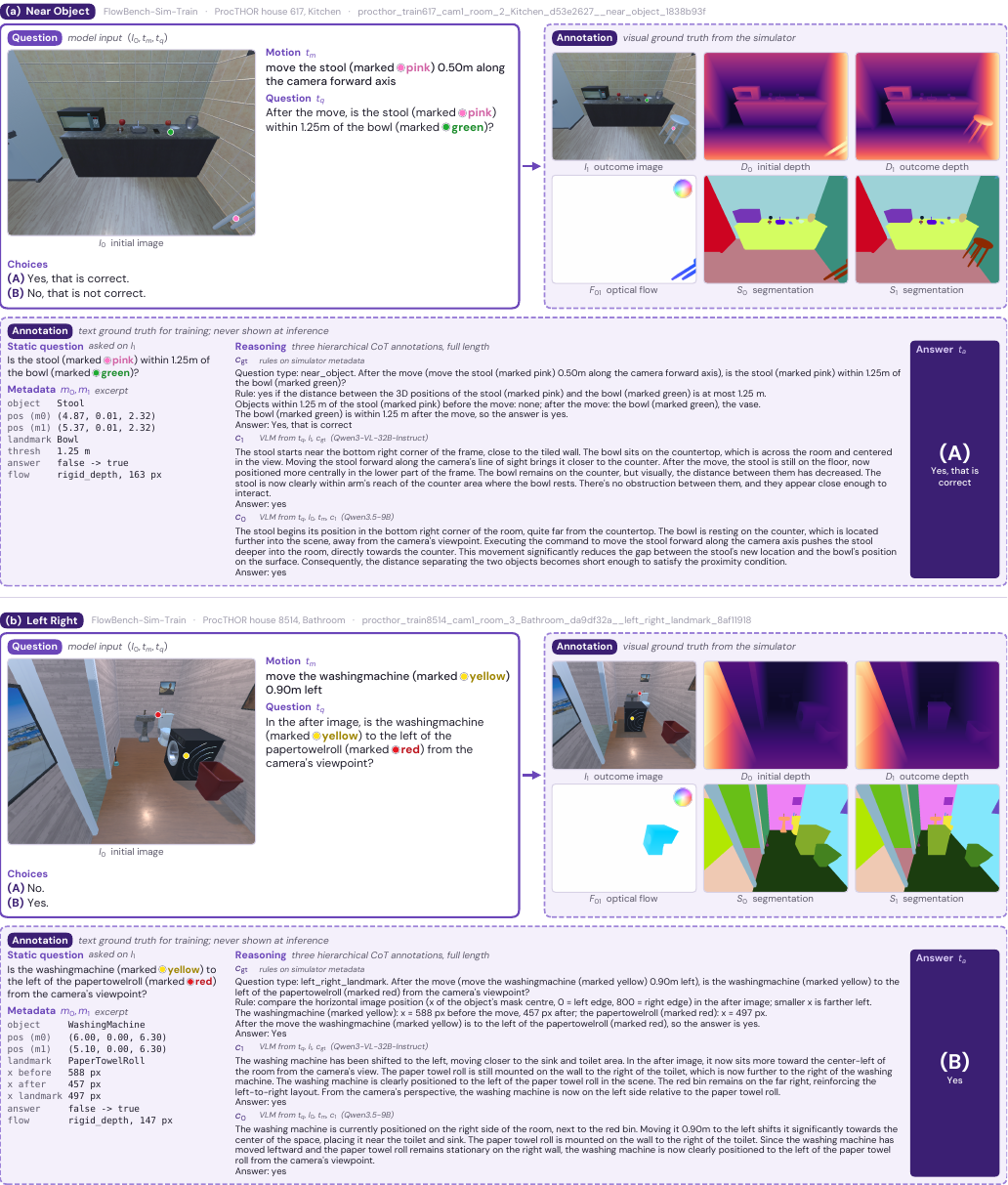}
\caption{Visualization of FlowBench-Sim-Train samples: (a) Near Object and (b) Left Right.}
\label{fig:vis_sim_4}
\end{figure*}

\begin{figure*}[p]
\centering
\includegraphics[width=\textwidth,height=\dimexpr\textheight-3\baselineskip\relax,keepaspectratio]{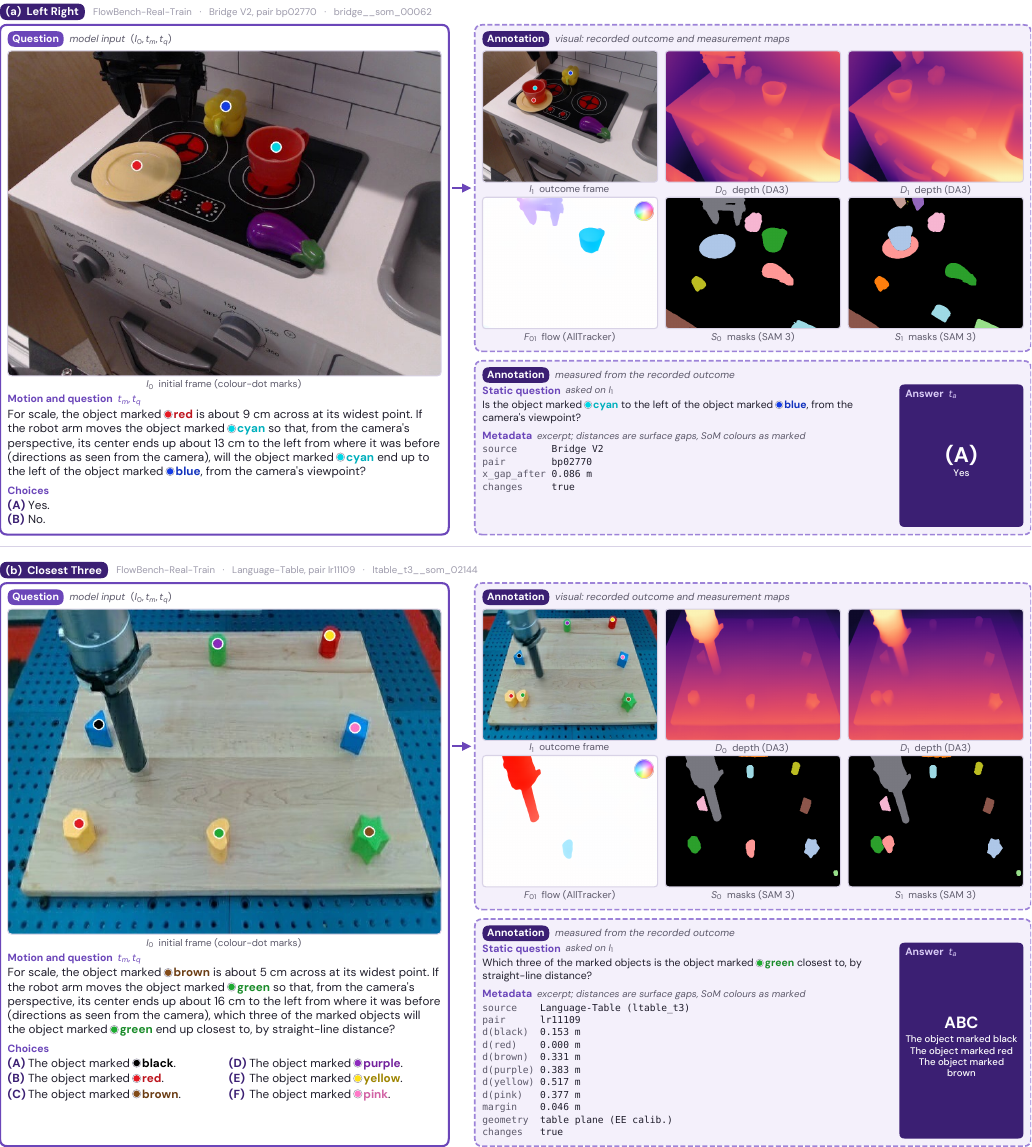}
\caption{Visualization of FlowBench-Real-Train samples: (a) Left Right from Bridge V2 and (b) Closest Three from Language-Table.}
\label{fig:vis_real_1}
\end{figure*}

\begin{figure*}[p]
\centering
\includegraphics[width=\textwidth,height=\dimexpr\textheight-3\baselineskip\relax,keepaspectratio]{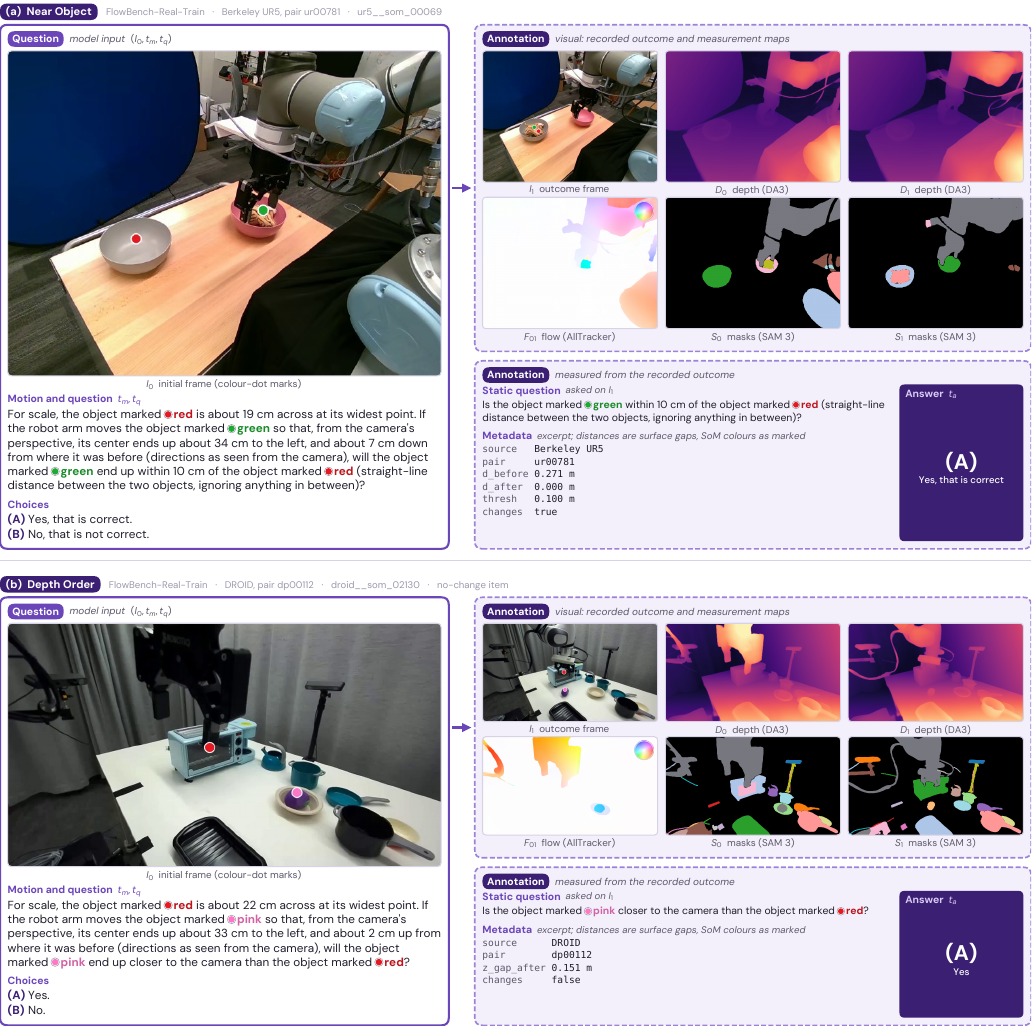}
\caption{Visualization of FlowBench-Real-Train samples: (a) Near Object from Berkeley UR5 and (b) Depth Order from DROID, a no-change question.}
\label{fig:vis_real_2}
\end{figure*}

\begin{figure*}[p]
\centering
\includegraphics[width=\textwidth,height=\dimexpr\textheight-3\baselineskip\relax,keepaspectratio]{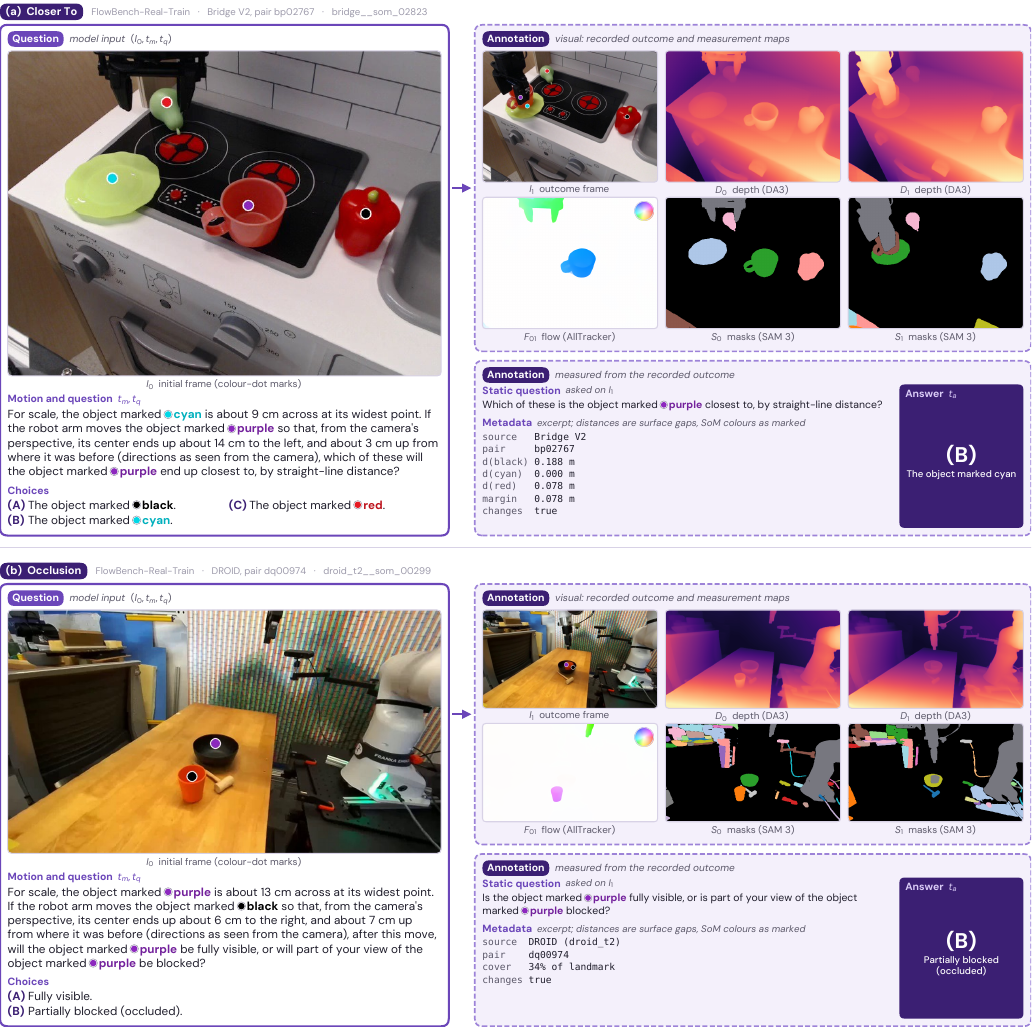}
\caption{Visualization of FlowBench-Real-Train samples: (a) Closer To from Bridge V2 and (b) Occlusion from DROID.}
\label{fig:vis_real_3}
\end{figure*}

\begin{figure*}[p]
\centering
\includegraphics[width=\textwidth,height=\dimexpr\textheight-3\baselineskip\relax,keepaspectratio]{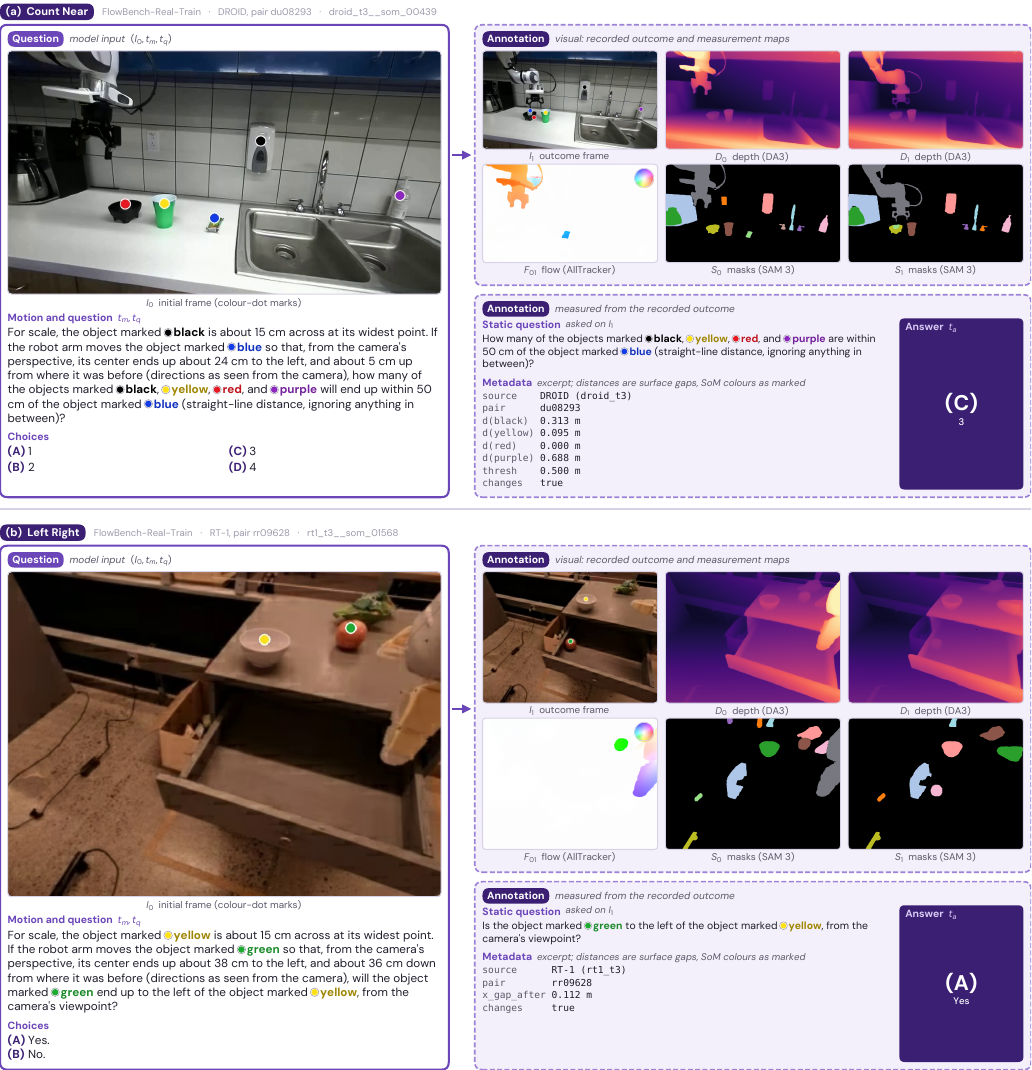}
\caption{Visualization of FlowBench-Real-Train samples: (a) Count Near from DROID and (b) Left Right from RT-1.}
\label{fig:vis_real_4}
\end{figure*}

We visualize two examples per page from the training splits: \cref{fig:vis_sim_1,fig:vis_sim_2,fig:vis_sim_3,fig:vis_sim_4} cover all eight FlowBench-Sim categories, and \cref{fig:vis_real_1,fig:vis_real_2,fig:vis_real_3,fig:vis_real_4} cover all seven FlowBench-Real categories and all five robot datasets.

\noindent\textbf{FlowBench-Sim.}
Each example shows the model input $(I_0,t_m,t_q)$ with its answer choices, followed by the annotations, which are never shown at inference. The visual ground truth is rendered by the simulator: the outcome image $I_1$, the depths $D_0,D_1$, the optical flow $F_{01}$, and the segmentations $S_0,S_1$. The text ground truth consists of the static question asked on $I_1$, an excerpt of the object metadata $m_0,m_1$, the three hierarchical reasoning annotations, and the answer $t_a$. The reasoning annotations are $c_{\mathrm{gt}}$, computed by rules on the simulator metadata; $c_1$, generated by Qwen3-VL-32B-Instruct from $(t_q,I_1,c_{\mathrm{gt}})$; and $c_0$, generated by Qwen3.5-9B from $(t_q,I_0,t_m,c_1)$.

\noindent\textbf{FlowBench-Real.}
Each example shows the model input: the initial frame $I_0$ with colored-dot marks, and the motion and question $(t_m,t_q)$, which state a visible object's width as a scale reference. The annotations are measured from the recorded video: the outcome frame $I_1$, the depths $D_0,D_1$ from Depth Anything~3, the flow $F_{01}$ from AllTracker, and the masks $S_0,S_1$ from SAM~3, together with the static question asked on $I_1$, a metadata excerpt with the measured surface gaps, and the answer $t_a$. Real examples have no reasoning annotations. \cref{fig:vis_real_2}(b) shows a no-change question, whose answer is the same before and after the hypothetical move (\cref{app:real_nochange}).

\clearpage

\clearpage
\section{Background on $\Psi$-0.6}
\label{app:psi}

\noindent\textbf{PSI.}
$\Psi$-0.6~\cite{psi} builds on Probabilistic Structure Integration (PSI)~\cite{kotar_world_2025}, which learns promptable world models from video in a cycle: a random-access autoregressive model predicts any subset of visual variables from any other, structures such as flow, depth, and segments are extracted from it zero-shot, and these are integrated back as new token types. PSI extracts state-of-the-art optical flow, self-supervised depth, and segmentation; \mbox{$\Psi$-0.6} extends it to language.

\noindent\textbf{Lineage.}
Building on counterfactual world modeling~\cite{bear_unifying_2023}, models in this line achieve state-of-the-art results on physical dynamics prediction~\cite{venkatesh_understanding_2024} and on novel-view synthesis and 3D object manipulation~\cite{lee_3d_2025,lee_unified_2026}. They match or exceed state-of-the-art motion estimation without flow supervision~\cite{stojanov_self-supervised_2025,kim_taming_2025}, even when trained on a single child's visual experience~\cite{aw_zero-shot_2026}. They also segment objects better than SAM~\cite{venkatesh_discovering_2025}, manipulate objects in 3D~\cite{venkatesh_physical_2026}, and simulate 3D transformations~\cite{lee_perceptual_2026}.

\noindent\textbf{Preliminaries.}
Each $16{\times}16$ RGB, depth, or flow patch is quantized independently and enters the sequence after a pointer token giving its modality and spatiotemporal location. A prompt can therefore condition on any subset of patches and generate the rest in any order, so a task is a choice of what to give and what to request: $(I_0,t_m)\to F_{01}$ is text-to-flow prediction, and imagine-then-answer chains $(I_0,t_m)\to\hat{F}_{01}\to\hat{I}_1\to t_a$. Motion-first decoding generates the patches most likely to move first (Suppl.~\ref{app:wm}); outputs are sampled, and classifier-free guidance~\cite{ho2022cfg} strengthens the motion command.

\end{document}